\documentclass{article} 
\usepackage[final]{colm2026_conference}

\usepackage{microtype}
\usepackage{hyperref}
\usepackage{url}
\usepackage{booktabs}
\usepackage{amsmath}
\usepackage{graphicx}
\usepackage[many,most,skins,theorems]{tcolorbox}
\usepackage{algorithm}
\usepackage{algpseudocode}
\usepackage{wrapfig}
\usepackage{xcolor}
\usepackage{algorithmicx}
\usepackage{multirow} 

\usepackage{amssymb}
\usepackage{mathtools}
\usepackage{amsthm}
\usepackage[table]{xcolor}
\usepackage{fontawesome5}
\usepackage{bbding}
\usepackage{tcolorbox}
\usepackage{listings}
\usepackage{threeparttable}
\usepackage{xspace}
\usepackage{float}  
\usepackage{placeins}

\newcommand{\toolrzero}{\textit{Tool-R0}\xspace}

\definecolor{acadNavy}{RGB}{40,70,140}      
\definecolor{acadMustard}{RGB}{180,140,30} 
\newcommand{\Generator}{\textcolor{acadMustard}{\textbf{\texttt{Generator}}}}
\newcommand{\Solver}{\textcolor{acadNavy}{\textbf{\texttt{Solver}}}}

\NewDocumentCommand{\heng}
{ mO{} }{\textcolor{red}{\textsuperscript{\textit{Heng}}\textsf{\textbf{\small[#1]}}}}

\definecolor{tzBlueHeader}{RGB}{78,160,205}
\definecolor{tzBlueHeader2}{RGB}{105,185,225}
\definecolor{tzBlueBorder}{RGB}{115,190,225}
\definecolor{tzBlueFill}{RGB}{232,246,252}

\definecolor{rqBlueBorder}{HTML}{6AADE4}

\usepackage{xcolor}

\definecolor{acadGreen}{RGB}{40,140,90}
\colorlet{acadNavyBg}{tzBlueFill!60}
\colorlet{acadMustardBg}{acadMustard!6}
\colorlet{acadGreenBg}{acadGreen!7}

\newcommand{\AlgSection}[2]{%
  \Statex \colorbox{#1}{\parbox{\dimexpr\linewidth-2\fboxsep\relax}{#2}}%
}

\newcommand{\researchq}[2]{%
\noindent{\fontsize{10.50}{13}\selectfont\textbf{Research Question #1: #2}}
}

\usepackage{cleveref}

\crefname{section}{Sec.}{Sec.}
\crefname{theorem}{Theorem}{Theorems}
\crefname{corollary}{Corollary}{Corollaries}
\crefname{lemma}{Lemma}{Lemmas}
\crefname{equation}{Eq.}{Eq.}
\crefname{proposition}{Proposition}{Propositions}
\crefname{claim}{Claim}{Claims}
\crefname{remark}{Remark}{Remarks}
\crefname{observation}{Observation}{Observations}
\crefname{assumption}{Assumption}{Assumptions}
\crefname{template}{Template}{Template}
\crefname{definition}{Definition}{Definitions}
\crefname{appendix}{App.}{Apps.}
\crefname{algorithm}{Algorithm}{Algorithms}
\crefname{figure}{Fig.}{Fig.}
\crefname{table}{Table}{Tables}
\crefname{property}{Property}{Properties}
\crefname{line}{Line}{Lines}

\DeclareTextCommand{\textquotedbl}{OT1}{\char`\"}

\lstdefinestyle{jsonTiny}{
  basicstyle=\ttfamily\scriptsize,
  breaklines=true,
  breakindent=0pt,
  columns=fullflexible,
  keepspaces=true,
  showstringspaces=false,
  upquote=true,
  frame=none
}

\tcbuselibrary{skins,breakable}

\newtcolorbox{researchstatement}[1][]{%
  enhanced,
  breakable,
  colback=rqBlueBg!18,
  colframe=rqBlueBorder!55,
  boxrule=1.25pt,
  arc=3pt,
  left=2mm,right=2mm,top=2mm,bottom=1.3mm,
  before skip=10pt, after skip=10pt,
  title={Research Statement},
  colbacktitle=rqBlueBorder!75,
  coltitle=black!95,
  fonttitle=\bfseries\small,
  attach boxed title to top left={yshift=-1.2mm, xshift=2mm},
  boxed title style={
    enhanced,
    arc=3pt,
    top=0.5mm, bottom=0.5mm, left=1mm, right=1mm,
    boxrule=0pt,
    interior engine=empty,
  },
  #1
}

\newtcolorbox{block}[1][]{%
  enhanced,
  breakable,
  colback=white,
  colframe=black!85,
  boxrule=1.4pt,
  arc=7pt,
  left=4mm,right=4mm,top=4mm,bottom=3mm,
  before skip=10pt, after skip=10pt,
  #1
}

\usepackage{lineno}

\definecolor{coco1}{HTML}{D9E4EC}
\definecolor{coco2}{HTML}{B7CFDC}
\definecolor{coco3}{HTML}{6AABD2}
\definecolor{coco4}{HTML}{385E72}
\hypersetup{colorlinks=true, linkcolor=coco3, filecolor=coco4, urlcolor=coco3, citecolor=coco3}

\definecolor{mylightgray}{RGB}{240,240,240}
\definecolor{rqBlueBg}{HTML}{EAF4FF}

\definecolor{TagGray}{RGB}{64,64,64}        
\definecolor{ToolGray}{RGB}{90,90,90}       

\newcommand{\thinkS}{\textcolor{TagGray}{$\texttt{<think>}$}}

\newcommand{\questionS}{\textcolor{TagGray}{$\texttt{<question>}$}}

\newcommand{\menuS}{\textcolor{TagGray}{$\texttt{<available\_tools>}$}}

\newcommand{\answerS}{\textcolor{TagGray}{$\texttt{<tool\_call\_answer>}$}}

\newcommand{\highlightpink}[1]{%
\tikz[baseline=(char.base)]{%
\node[%
shape=rectangle,%
rounded corners=2pt,%
draw=mylightgray,%
fill=mylightgray,%
inner sep=2pt,%
text width=0.95cm,%
align=center,%
text height=1.3ex,%
text depth=.1ex%
] (char) {#1};%
}%
}

\usepackage{enumitem}
\newenvironment{itemize*}%
 {\leftmargini=20pt\begin{itemize}%
  \setlength{\itemsep}{3pt}%
  \setlength{\parskip}{0pt}%
  }%
 {\end{itemize}} 
\newenvironment{enumerate*}%
 {\begin{enumerate}%
  \setlength{\itemsep}{0pt}%
  \setlength{\parskip}{0pt}}%
 {\end{enumerate}}

\newcommand{\hlbox}[1]{%
  \begingroup
  \setlength{\fboxsep}{1pt}
  \colorbox{rqBlueBg!50}{#1}%
  \endgroup
}

\tcbset{
  aibox/.style={
    width=\linewidth,
    top=7pt,
    bottom=2pt,
    colback=rqBlueBg,     
    colframe=black,
    colbacktitle=black,
    enhanced,
    center,
    attach boxed title to top left={yshift=-0.1in,xshift=0.15in},
    boxed title style={boxrule=0pt,colframe=white,},
  }
}
\newtcolorbox{AIbox}[2][]{aibox,title=#2,#1}

\newcounter{takeaway}
\newtcolorbox{takeaway}[1][]{
  aibox,
  colback=rqBlueBg,
  title={\stepcounter{takeaway}Takeaway \thetakeaway},
  #1
}

\NewDocumentCommand{\cheng}
{ mO{} }{\textcolor{orange}{\textsuperscript{\textit{Cheng}}\textsf{\textbf{\small[#1]}}}}

\title{\scalebox{1.20}{\textit{Tool-R0}:}\\Self-Evolving LLM Agents for Tool-Learning from Zero Data}

\author{%
Emre Can Acikgoz\textsuperscript{$1$}, \; 
Cheng Qian\textsuperscript{$1$}, \; 
Jonas Hübotter\textsuperscript{$2$}, \\
\bfseries Heng Ji\textsuperscript{$1$}, \; 
Dilek Hakkani-Tür\textsuperscript{$1$}, \; 
Gokhan Tur\textsuperscript{$1$} \\
\textsuperscript{$1$}UIUC\; 
\textsuperscript{$2$}ETH Zurich
}

\begin{document}

\ifcolmsubmission
\linenumbers
\fi

\maketitle

\begin{abstract}

Large language models (LLMs) are becoming the foundation for autonomous agents that can use tools to solve complex tasks. 
Reinforcement learning (RL) has emerged as a common approach for injecting such agentic capabilities, but typically under tightly controlled training setups.
It often depends on carefully constructed task–solution pairs and substantial human supervision, which creates a fundamental obstacle to open-ended self-evolution toward superintelligent systems.
In this paper, we propose \toolrzero framework for training general purpose tool-calling agents from scratch with self-play RL, \textit{under a zero-data assumption}.
Initialized from the same base LLM, \toolrzero co-evolves a \Generator\ and a \Solver\ with complementary rewards: one proposes targeted challenging tasks at the other's competence frontier and the other learns to solve them with real-world tool calls. 
This creates a self-evolving cycle that requires no pre-existing tasks or datasets.
Evaluation on different tool-use benchmarks show that \toolrzero\ yields \highlightpink{\textbf{92.5\%}}\ relative improvement over the base model and surpasses fully supervised tool-calling baselines under the same setting. 
Our work further provides empirical insights into self-play LLM agents by analyzing co-evolution, curriculum dynamics, and scaling behavior.


\end{abstract}

\definecolor{academicblue}{RGB}{0,70,140}
\definecolor{linkgray}{RGB}{40,40,40}
\definecolor{acnavy}{RGB}{0,38,77}
\definecolor{darkernavy}{RGB}{0,30,60}
\definecolor{darkestnavy}{RGB}{0,15,35}

\vspace{-3mm}
\begin{center}
\small

\newcommand{\logoh}{1.35em}

\href{https://github.com/emrecanacikgoz/Tool-R0}{
\raisebox{-0.2\height}{\includegraphics[height=\logoh]{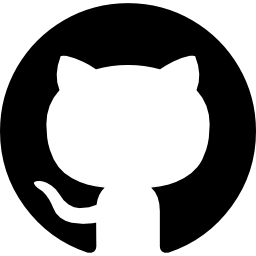}}
\hspace{0.35em}\textcolor{darkestnavy}{\textbf{Code}}
}
\quad
\href{https://emrecanacikgoz.github.io/Tool-R0}{
\raisebox{-0.2\height}{\includegraphics[height=\logoh]{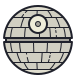}}
\hspace{0.35em}\textcolor{darkestnavy}{\textbf{Project Page}}
}
\quad
\href{https://huggingface.co/collections/emrecanacikgoz/tool-r0}{
\raisebox{-0.2\height}{\includegraphics[height=\logoh]{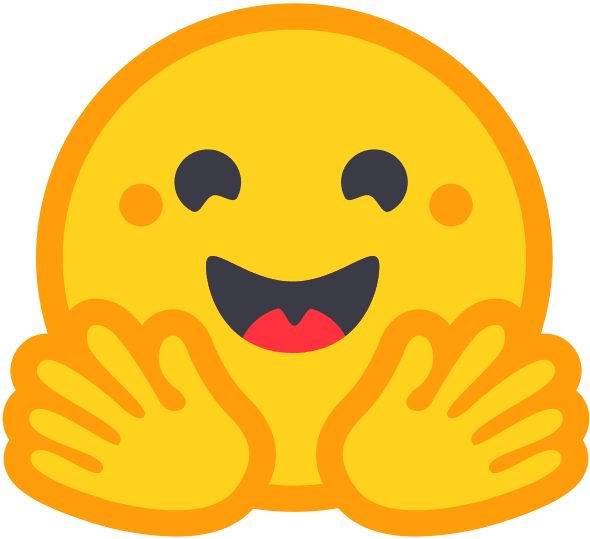}}
\hspace{0.35em}\textcolor{darkestnavy}{\textbf{Models}}
}
\quad
\href{https://api.wandb.ai/links/acikgoz2-university-of-illinois-urbana-champaign/olowdrg5}{
\raisebox{-0.2\height}{\includegraphics[height=\logoh]{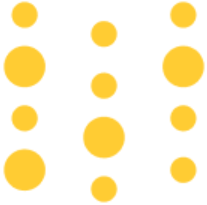}}
\hspace{0.35em}\textcolor{darkestnavy}{\textbf{Logs}}
}

\end{center}
\vspace{-2mm}

\section{Introduction}
\vspace{-2mm}
Large language models (LLMs) have demonstrated remarkable capabilities in complex reasoning tasks~\citep{jaech2024openaio1, guo2025deepseekr1, team2025kimi}. 
A key driver of this progress is reinforcement learning with verifiable rewards (RLVR), which has emerged as a powerful post-training paradigm for improving reasoning performance~\citep{lambert2025tulu}.
These advances position LLMs as promising language agents that can reason and act in real-world environments to accomplish goals by interacting with external tools~\citep{yao2023react, qu2025toolsurvey}.
To teach LLMs with these agentic capabilities, large-scale RLVR has become the focal point~\citep{zhang2025agenticrl-survey, qian2025toolrl, dong2026agenticpolicy}.

\begin{wrapfigure}{l}{0.42\linewidth}
\centering
\vspace{-4mm}
\includegraphics[width=\linewidth]{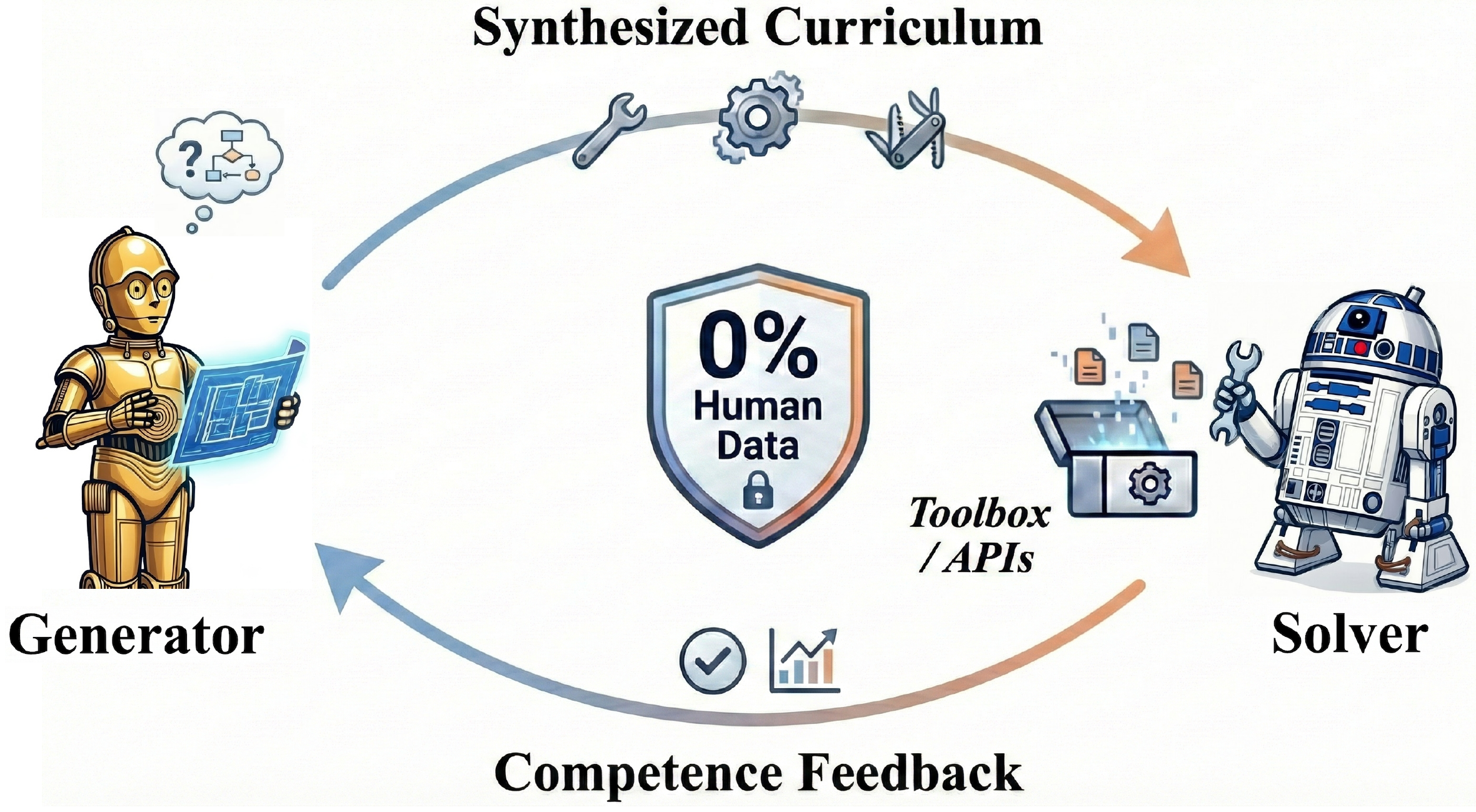}
\vspace{-8mm}
\caption{\toolrzero\ self-evolution loop.}
\vspace{-3mm}
\label{fig:selfplay-abs}
\end{wrapfigure}

Prior work on LLM agents has proposed different methods to improve their tool-calling skills over vast task datasets~\citep{zhang2025xlam, liu2025toolace, acikgoz2025coalm, qian2025toolrl}.
However, curating such datasets through human annotation is inherently labor-intensive and fundamentally unscalable~\citep{huang2023llmsselfimprove, weizge2024selfrewarding}.
As models advance, constructing large and high-quality datasets becomes increasingly unsustainable, creating a scalability bottleneck similar to that observed in LLM pretraining paradigm~\citep{neurips2024testoftime, silver2025welcome}.
Consequently, reliance on human supervision poses a fundamental scalability bottleneck, limiting the ability of LLM agents to achieve self-evolving capabilities that a superintelligent AI would require~\citep{zhang2025darwin, gao2025selfevolve-survey}.
This naturally raises the following question: \textit{Can we empower weak LLMs to improve themselves without requiring human data?}

Some of the most striking examples of superhuman AI emerge from self-play using RL in game environments~\citep{silver2016alphago, Silver2017alphagozero, brown2022cicero}.
Recently, interest in self-play has re-emerged with the success of LLMs trained via RL; where several works have applied it to reasoning tasks like math~\citep{chen2024spin, cheng2024spag, chen2025spc, zhao2025absolute, liu2025spiral, huang2025r-zero, liu2025spice}. 
Despite this progress, these methods do not extend self-play beyond abstract reasoning to tool-using agents in real-world environments.
This further motivates whether self-play alone can incentivize LLMs with tool-calling skills across different domains under a zero-data assumption.
Recent efforts such as Agent0~\citep{xia2025agent0} and Dr. Zero~\citep{yue2026drzero} have begun to explore this direction. 
However, their focus remains narrow, either constrained to code tools for math problem-solving or search tools in question answering settings. 
In contrast, we focus on \textbf{how to train general-purpose tool-calling agents that can self-evolve base LLMs with zero external data across domains}.

We propose \toolrzero, a self-play RL framework for training tool-calling agents that can self-evolve with zero human data (\cref{fig:selfplay-abs}).
In \toolrzero, a single base model is initialized with two different roles: a \Generator\ and a \Solver\ that are trained independently but co-evolve through complementary reward signals. 
The \Generator\ is rewarded for generating challenging tasks aligned with the \Solver’s evolving capabilities, while the \Solver\ is trained to solve them with outcome-based rewards.
To enable effective co-evolution, we design a difficulty-guided reward based on the frozen \Solver's answer uncertainty, prioritizing tasks at its competence frontier: \textbf{hard enough to teach and easy enough to verify.}
We also follow a grounded task synthesis to ensure adaptive and controllable task generation, which conditions the \Generator\ through annotation-free task configurations to prevent mode collapse in free-form generation.
Since effective RL requires training data aligned with the model’s current capabilities, we further order the constructed dataset by \Solver\ answer consistency and progressively expose it to harder problems in each batch.
Together, this self-evolving cycle significantly improves tool-calling performance over the base LLM and outperforms other supervised baselines across five different tool-calling benchmarks.
We summarize the core insights derived from our study as follows:
\begin{itemize}[topsep=-1.5pt, leftmargin=10pt, itemsep=-0pt]

    \item \textbf{Self-play can incentivize complex tool-calling skills from zero data.}
    Starting from weak priors, \toolrzero\ yields consistent gains across model scales, architectures, and benchmarks, showing that self-play RL alone is sufficient to teach complex tool-calling capabilities.
    
    \item \textbf{Adaptive self-generated curricula can surpass static human supervision.}
    Agents trained through self-play generalize better than those trained on fixed human-distribution datasets, demonstrating that LLMs can leverage themselves to identify and address their own capability gaps more effectively than fixed expert-designed curricula.
    
    \item \textbf{Role separation is necessary to avoid learning conflicts in self-play.}
    We observe that training \Generator\ and \Solver\ with separate parameters is essential, particularly when both roles operate over high-entropy action spaces with different reward objectives.

    \item \textbf{Co-evolution with difficulty-aware rewards drives learning.} 
    We find that effective learning requires a dynamic \Generator\ guided by smooth difficulty-aware rewards, since static generation or miscalibrated difficulty fails to sustain tool-use improvement.
\end{itemize}
\vspace{-0.1cm}
\begin{AIbox}{Overview of contributions}
    \begin{itemize}[leftmargin=0em]
        \item \textbf{(Algorithmic)} The primary contribution of this paper is \toolrzero, an adaptive self-play RL framework, where a \Generator\ and a \Solver\ co-evolve to build general-purpose tool agents with zero data from scratch.
        
        \item \textbf{(Performance)} \toolrzero\ consistently improves base models across architectures, scales, and benchmarks, and surpasses fully supervised tool agents, showing that LLMs can self-evolve into general tool-calling agents in a data-free setting.
        
        \item \textbf{(Analysis)} We conduct a series of experiments to analyze the key mechanisms of zero-data self-play for tool learning, including: role asymmetry, curriculum difficulty, adversarial co-evolution, self-play as mid-training stage, and scaling behaviors.
        
        \item \textbf{(Implementation)} We provide a modular and plug-and-play repository of \toolrzero\ to support future research on self-evolving agents.

    \end{itemize}
\end{AIbox}

\section{Related Work}
\vspace{-2mm}

\noindent\textbf{Tool-Learning with LLMs.}
Tool-learning has emerged as a promising paradigm for extending LLMs capabilities beyond parametric knowledge with external tools~\citep{qu2025toolsurvey}.
It requires LLMs to perform structured reasoning over available tool schemas by selecting the right function, grounding arguments from context, and composing multi-step calls to accomplish goals~\citep{yao2023react, schick2023toolformer, patil2024gorilla}.
Several benchmarks have been proposed to systematically assess these capabilities across domains, task types, and tool sets~\citep{tang2023toolalpaca, srinivasan2023nexusraven, li2023apibank, wu2024sealtool}.
Building on this foundation, subsequent work has focused on constructing high-quality datasets and fine-tuning LLMs on such data to integrate more complex tool-calling skills~\citep{zeng2024agenttuning, chen2024agentflan, zhang2025xlam, liu2025toolace, lin2025hammer, acikgoz2025coalm}, including applying RLVR to further enhance tool-integrated reasoning (TIR)~\citep{jin2025searchr1, feng2025retool, qian2025toolrl}.
However, their dependence on human supervision introduces limitations: (1) static human-curated data leads to distribution shifts that fail to capture the evolving needs of agents during on-policy learning and (2) limits the model's capabilities to human upper-bound.
To the best of our knowledge, \toolrzero is the first work to leverage self-play RL to \textit{train domain-agnostic agents entirely from scratch without any human data}, matching the performance of supervised ones, and demonstrating the promise of absolute zero paradigm in LLM agents~\citep{zhao2025absolute}.

\noindent\textbf{Self-Evolution through Self-Play.}
Self-play has historically driven superhuman performance in closed-loop games~\citep{tesauro1995tdgammon, silver2016alphago, brown2022cicero}. 
Recently, interest in self-play has re-emerged with the success of LLMs trained via RLVR; where previous work extends self-play to open-domain settings like reasoning~\citep{chen2024spin, cheng2024spag, zhao2025absolute, fang2025serl, chen2025spc, liu2025spiral, huang2025r-zero, liu2025spice}. 
These works focus exclusively on abstract reasoning tasks like math, and do not address tool-using agents for real-world applications.
More recent studies have begun exploring agentic self-play; Agent0~\citep{xia2025agent0} applies self-play with Python coding tools and Dr. Zero~\citep{yue2026drzero} studies zero-data learning for search-centric question answering. 
However, these methods remain narrowly scoped in terms of tool-learning, relying on a single tool type and limited action spaces like question answering. 
\toolrzero\ advances this line of work with a \textit{general-purpose} co-evolutionary framework where LLMs autonomously learn open-ended tool-use across arbitrary domains driven by user needs.
We defer a more comprehensive background in Appendix \ref{app: related-work-extened}.

\begin{figure}[!t]
    \centering 
    \vspace{-9mm}
    \includegraphics[width=\linewidth]{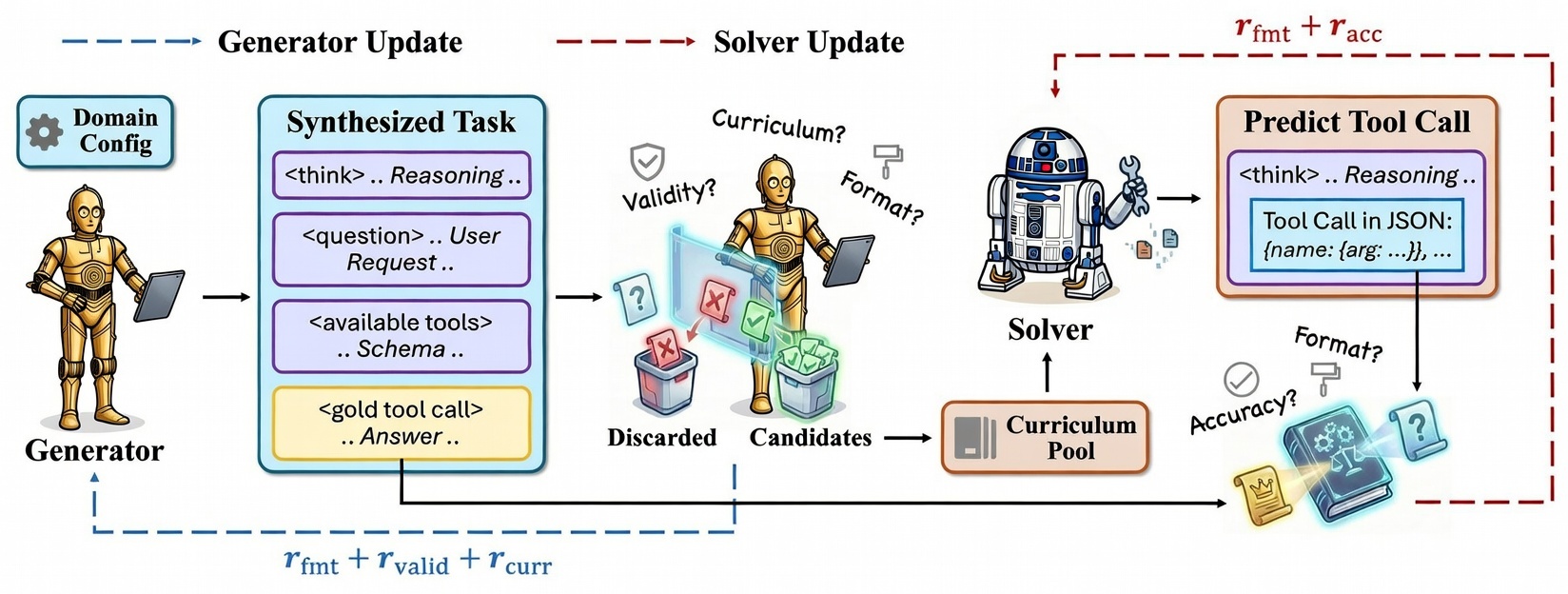}
    \vspace{-9mm}
    \caption{
    \textbf{\toolrzero\ Self-Evolution Framework.}
    A base LLM is initialized into two roles: a \Generator\ and a \Solver.
    The \Generator\ synthesizes challenging tool-calling tasks (question, tool menu, and gold tool-call), targeting the frozen \Solver's competence frontier with designed rewards ($r_{\text{fmt}} + r_{\text{valid}} + r_{\text{curr}}$).
    Generated tasks are filtered and ranked easy-to-hard into a curriculum pool.
    The \Solver\ trains on this curated data to predict tool calls ($r_{\text{fmt}} + r_{\text{acc}}$), completing a self-evolving cycle that requires no pre-existing human datasets.
    }
    \vspace{-5mm}
    \label{fig:pipeline}
\end{figure}

\vspace{-2mm}
\section{Method}
\label{sec:method}
\vspace{-2mm}

\noindent\textbf{Overview.\;\;} 
We propose \toolrzero, a zero-data self-play framework for training general-purpose tool-calling agents via dual-RL between two components: a \Generator\ that synthesizes verifiable tool-use tasks, and a \Solver\ that learns to solve them (\Cref{fig:pipeline}), both initialized from the same base LLM ($\pi$). 
Training proceeds for $K$ self-play iterations. 
At each iteration, given domain configurations (\Cref{subsec:task-config}), we first train \Generator\ ($\pi_\theta$) to produce grounded tool-calling tasks that adaptively target the \Solver’s evolving competence frontier (\Cref{subsec:method-generator}).
We then freeze the trained \Generator\  and use it to construct a high-quality dataset via deduplication, cross-verification, and difficulty-based curriculum ordering (\Cref{subsec:dataset-construction}). 
Finally, the \Solver\ ($\pi_\phi$) is trained on this curated dataset (\Cref{subsec:method-solver}) and carried forward to the next iteration.
This co-evolutionary process drives the \Generator\ to propose progressively harder yet solvable tasks across iterations, while the \Solver\ continuously adapts to an expanding tool-use curriculum; without any human-authored data (full pseudocode in Appendix~\ref{app:algortihm}).

\subsection{Grounded Task Specification.}
\label{subsec:task-config}
A key challenge in zero-data self-play is controllable and diverse task generations: when the \Generator\ is prompted only with generic instructions to produce diverse tasks, the underlying language model concentrates probability mass on a small set of high-likelihood patterns, resulting in mode-collapsed generations~\citep{liu2025spice}. 
To prevent this while preserving domain-agnostic adaptability, \toolrzero\ grounds task generation using lightweight task specifications $s=(d,c,m,n)$ where $d$ denotes the task domain (\Cref{fig:domain-config}), $c$ the interaction context type, $m$ the number of available tools, and $n$ the number of gold tool-calls as answer.
At each training step, specifications are dynamically sampled from a user-defined weighted distribution and injected into the \Generator's prompt (\Cref{fig:generator-prompt-template}) as a form of meta-prompting resulting with conditioned generation distribution: $q \sim \pi_\theta(\cdot \mid s)$, where constrained generation improves specificity, diversity, and robustness to hallucination.
Formal definitions, sampling strategies, and hyperparameters are provided in Appendix \ref{app:task-specification}.

\subsection{Generator Training}
\label{subsec:method-generator}
\toolrzero\ begins by training a \Generator\ policy $\pi_\theta$ with Group Relative Policy Optimization (GRPO)~\citep{shao2024deepseekmath} to synthesize challenging tool-calling tasks from scratch, using no external data and only the prompt shown in \Cref{fig:generator-prompt-template} together with task specifications $s$.
Each generated sample consists of: (i) a user request, (ii) an explicit tool menu, and (iii) a gold tool-call, within its private reasoning.
Concretely, the \Generator\ is asked to output \emph{exactly} four tagged blocks: \thinkS, \questionS, \menuS, and \answerS. 
The \menuS\ block is a JSON list of tool specifications (e.g., name, description, and a JSON schema for parameters),
and \answerS\ is a JSON list of tool calls with flat primitive arguments only, so that tool menus and calls are JSON-verifiable; enabling execution-based feedback rather than non-verifiable natural-language supervision (see \cref{fig:gen-evolution-travel-hard} for a generated example).

\subsubsection{Generator Reward Design} 
Training the \Generator\ in a zero-data regime requires rewards that simultaneously enforce a reliable and targeted learning signal for the \Generator.
The core of this process is designing reward functions that capture what makes a question both well-posed and useful for learning.
We therefore train the \Generator\ with a set of rewards that serve three complementary purposes:
(i) enforce a strict code verifiable output interface ($r_{\text{fmt}}$), 
(ii) guarantee internal consistency between the tool menu and the gold tool-call ($r_{\text{valid}}$), and 
(iii) induce an adaptive curriculum by targeting tasks that are neither trivial nor unsolvable for the current \Solver\ ($r_{\text{curr}}$).

\paragraph{Format Reward ($r_{\text{fmt}}$): Tags and Parseability.}
Since downstream learning and verification require extracting the tool menu and gold tool-calls, we first enforce strict structural compliance.
Let $x$ denote a \Generator\ completion and let $\mathbb{I}_{\text{tags}}(x)$ indicate that all required blocks can be extracted.
We reward tag completeness and well-formed JSON artifacts:
\begin{equation}
r_{\text{fmt}}(x)
=
\mathbb{I}_{\text{tags}}(x)
+
\mathbb{I}_{\text{tools-json}}(x)
+
\mathbb{I}_{\text{gold-json}}(x),
\label{eq:1}
\end{equation}
where $\mathbb{I}_{\text{tools-json}}(x)$ indicates that \menuS\ parses as a JSON list of tool specs, and $\mathbb{I}_{\text{gold-json}}(x)$ indicates that \answerS\ parses and normalizes into a canonical tool-call representation.
This component ensures that \Generator\ outputs are verifiable and executable, preventing reward hacking through malformed outputs.

\paragraph{Validity Reward ($r_{\text{valid}}$): Available Tools, Gold-Calls, and Value Grounding.}
Syntactic correctness alone does not guarantee that the proposed gold answer is even realizable using the provided tools.
We therefore enforce internal consistency between the tool menu, the gold tool asnwer, and the question itself to avoid action hallucination.
Let $\mathcal{T}$ denote the successfully parsed tool menu, let $c^\star=(n^\star,a^\star)$ be the normalized gold tool call with tool name $n^\star$ and flat argument map $a^\star$, and $q$ is the generated question\footnote{When \answerS\ contains a list of calls, we canonicalize calls into a normalized tool-call representation for verification.}.
We assign validity rewards based on three checks:

\begin{equation}
r_{\text{valid}}(x)
=
\lambda_{Menu}\,\mathbb{I}[n^\star \in \mathcal{T}]
+
\lambda_{Gold}\,\mathbb{I}[\operatorname{req}(n^\star)\subseteq \operatorname{keys}(a^\star)]
+
\lambda_{Value}\,\mathbb{I}[\operatorname{vals}(a^\star)\hookrightarrow q],
\label{eq:2}
\end{equation}
where the three terms respectively verify that the gold tool exists in the menu, that all schema-required parameters are present, and that every non-trivial argument value (excluding booleans and nulls) appears as a word-boundary match in $q$ ($\operatorname{vals}(a^\star) \hookrightarrow q$).

\paragraph{Curriculum Reward ($r_{\text{curr}}$): Difficulty \& Semantic Alignment.}
Beyond syntactic validity,
\begin{wrapfigure}{r}{0.46\linewidth}
\centering
\vspace{-4mm}
\includegraphics[width=\linewidth]{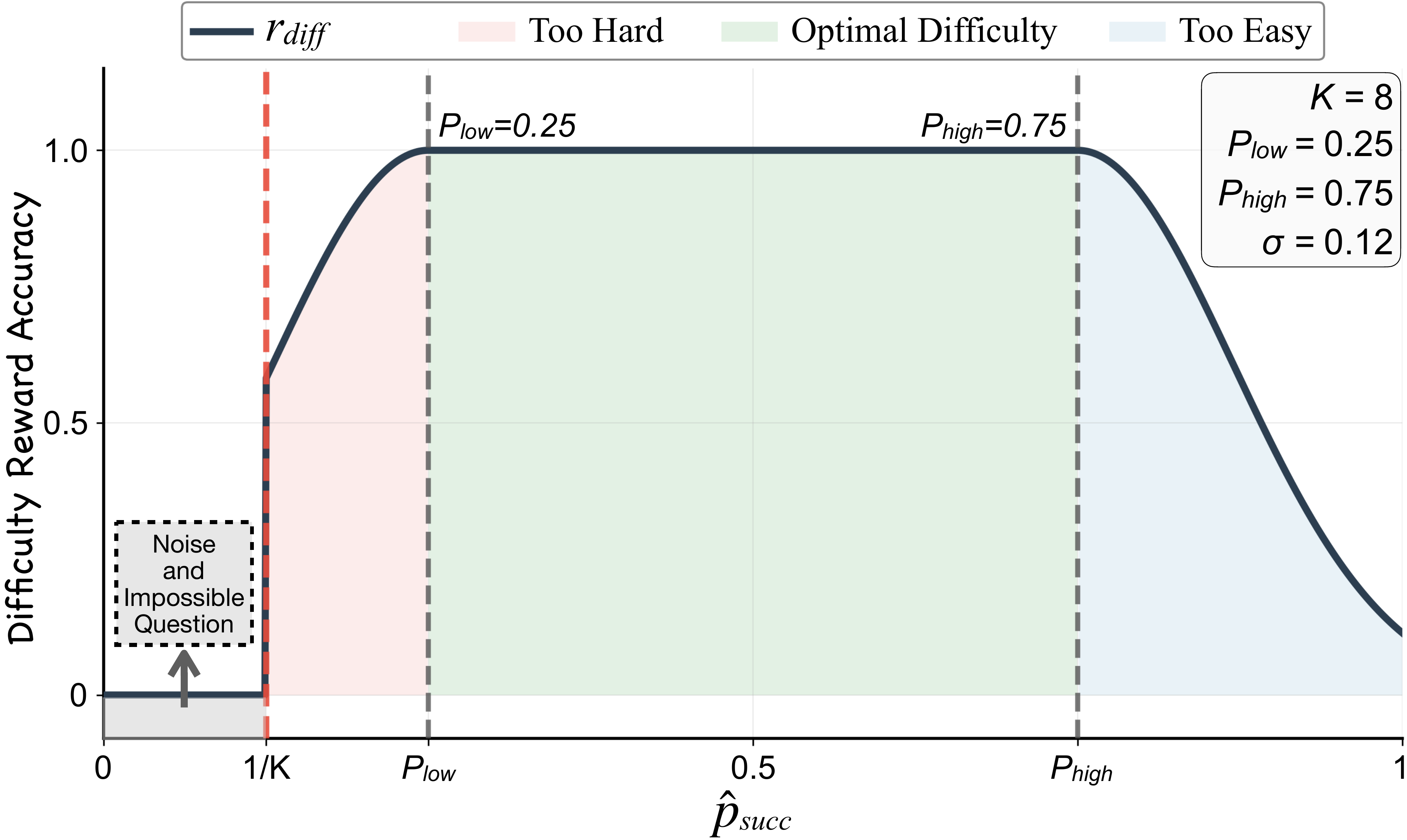}
\vspace{-8mm}
\caption{Difficulty reward ($r_{\text{diff}}$).}
\vspace{-8mm}
\label{fig:rdiff}
\end{wrapfigure}
we require generated tasks to be \emph{useful for learning}: they should (i) lie near the \Solver’s current competence frontier level and (ii) remain semantically faithful to the user request. We therefore define a curriculum-quality reward that combines \Solver-calibrated difficulty ($r_{diff}$) with a semantic coherence signal ($r_{sem}$).
Given a generated question and tool menu, we query the current \Solver\ $K$ times under Monte Carlo stochastic decoding to obtain predicted tool-calls $\{\hat{c}^{(k)}\}_{k=1}^K$ and estimate success against the gold tool-call $c^\star$:
\begin{equation}
\hat{p}_{\text{succ}}
=
\frac{1}{K}\sum_{k=1}^K \mathbb{I}\big[\hat{c}^{(k)} = c^\star\big].
\end{equation}
For solvable tasks, we assign maximal difficulty reward when $\hat{p}_{\text{succ}}$ lies in a target interval $[P_{\text{low}},P_{\text{high}}]$, with smooth decay outside:
\begin{equation}
r_{\text{diff}}(x)=
\begin{cases}
1, & \hat{p}_{\text{succ}}\in[P_{\text{low}},P_{\text{high}}],\\
\exp\!\left(-\frac{(\hat{p}_{\text{succ}}-P_{\text{low}})^2}{2\sigma^2}\right), & \hat{p}_{\text{succ}}<P_{\text{low}},\\
\exp\!\left(-\frac{(\hat{p}_{\text{succ}}-P_{\text{high}})^2}{2\sigma^2}\right), & \hat{p}_{\text{succ}}>P_{\text{high}},\\
0, & \hat{p}_{\text{succ}}<1/K.
\end{cases}
\label{eq:r_diff}
\end{equation}
which acts as a \textbf{band-pass signal filter} as illustrated in \Cref{fig:rdiff}. 
Here, the band-pass structure of $r_{\text{diff}}$ reflects three key insights about curriculum design. First, tasks with $\hat{p}_{\text{succ}} < 1/K$ (i.e., no solver sample matches the gold tool-call) are gated to zero reward, as they likely represent ill-posed, ambiguous, or unsolvable generations that provide no meaningful signal for \Solver. Second, rather than peaking at a single target probability, we reward a plateau around $\hat{p}_{\text{succ}} \approx 0.5$ ensuring tasks are neither trivially easy nor frustratingly hard. The plateau width absorbs the inherent noise in Monte Carlo estimation: with finite $K$ samples, a task near the true curriculum target should not be penalized due to sampling variance. 
Third, the Gaussian falloff outside the plateau provides a gradual decay in both directions: tasks that are too hard (low $\hat{p}_{\text{succ}}$) receive diminishing reward that encourages the Generator to produce slightly easier variants, while tasks that are too easy (high $\hat{p}_{\text{succ}}$) receive diminishing reward that encourages harder variants---both without the harsh cliffs that would destabilize training (see \cref{subsec:analysis} for ablation).

To ensure that the proposed gold tool-call meaningfully addresses the request, we introduce a semantic alignment reward ($r_{\text{sem}}$): the current \Solver\ scores the \Generator's output on a scale of $s\in{1,\ldots,5}$ for semantic coherence (\Cref{fig:semantic-alignment-prompt-template}), normalized to $[0,1]$ via $r_{\text{sem}}(x) = \frac{s(x)-1}{4}$.
We then define the combined curriculum-quality reward as:
\begin{equation}
r_{\text{curr}}(x) = r_{\text{diff}}(x) + r_{\text{sem}}(x),
\label{eq:4}
\end{equation}
which encourages the \Generator\ to propose tasks that are both appropriately challenging for the evolving \Solver\ and semantically well-posed.

\subsection{Solver Dataset Construction}
\label{subsec:dataset-construction}
After training, we freeze the \Generator\ and use it as a task synthesizer conditioned on the same control specifications. We first sample a large pool of candidate tasks and remove near-duplicates via canonicalized signatures derived from question–tool–call combinations. To increase the reliability of pseudo-labels, we cross-verify each candidate by sampling multiple predictions from \Solver\ and measuring agreement with the generated gold tool-call by retaining only tasks with consistent solutions; following the principle that reproducible answers provide more reliable supervision~\citep{huang2023llmsselfimprove, zuo2025ttrl}. From this verified pool, we estimate difficulty via pass@$K$ success rates and group tasks into easy, medium, and hard buckets. The final dataset is selected to preserve domain diversity while maintaining a balanced difficulty mix, organized as a curriculum progressing from easier to harder instances (see Appendix \ref{app:dataset-construction} for details).

\subsection{Solver Training}
\label{subsec:method-solver}

Our \Solver\ $\pi_\phi$ is trained to predict correct tool calls given a user query $q$ and tool menu $\mathcal{T}$ using a prompt template that elicits explicit reasoning (\Cref{fig:solver-prompt-template}), following prior work on reasoning-oriented RLVR~\citep{guo2025deepseekr1}. 
We train \Solver\ in TIR settings following ToolRL~\citep{qian2025toolrl}; it first generates reasoning within \thinkS\ tags, followed by predicted tool calls in \answerS\ tags. We use two different reward categories for our training: (i) a format reward $r_{fmt}$ and (ii) a dense accuracy reward $r_{\text{acc}}$.

\noindent\textbf{Format reward ($r_{fmt}$).}\;
To stabilize training and avoid unstructured outputs, we assign partial credit on parseability for our special tokens:
\begin{equation}
r_{fmt}(\hat{y})=
    \lambda_{tag}\cdot\mathbb{I}_{\text{tag}}
    +\lambda_{parse}\cdot\mathbb{I}_{\text{parse}}
    +\lambda_{norm}\cdot\mathbb{I}_{\text{norm}},
\end{equation}
where $\hat{y}$ is the model completion and normalization maps common tool-call variants into the canonical schema.

\paragraph{Accuracy Reward ($r_{\text{acc}}$).}
For parsed outputs, we decompose tool-call correctness into three components following prior work~\citep{qian2025toolrl}. 
Given predicted tool-calls $\hat{C}$ and gold tool-calls $C^\star=[c^\star_1,\ldots,c^\star_n]$, we match each gold tool-call to the best unused prediction (greedy maximum match).
For a matched pair $(\hat{c},c^\star)$, we compute three dense sub-scores:
(i) name match $s_{\text{name}}\in\{0,1\}$,
(ii) key overlap $s_{\text{key}}\in[0,1]$ measured by F1 over argument keys,
and (iii) value match $s_{\text{val}}\in[0,1]$ defined as the fraction of matching values over the intersection of keys, using a robust comparator (numeric coercion and whitespace-insensitive string matching).
The pair score is
\begin{equation}
s(\hat{c},c^\star)
=\lambda_{\text{name}}s_{\text{name}}
+\lambda_{\text{key}}s_{\text{key}}
+\lambda_{\text{val}}s_{\text{val}},
\end{equation}
We average matched scores over gold tool-calls to obtain a base accuracy $\bar{s}$.

To discourage spurious tool usage, we downweight over-prediction with a multiplicative penalty for extra calls: $r_{\text{acc}} = \bar{s}\cdot \frac{1}{1+\alpha\cdot \max(0,|\hat{C}|-|C^\star|)}$.
As the \Solver\ improves, its success statistics directly influence the difficulty reward of the \Generator\ (\Cref{eq:r_diff}), closing the self-play loop.
This co-evolutionary process drives \Generator\ to propose progressively harder yet solvable tasks, while the \Solver\ continuously adapts to the expanding tool-use curriculum; without any external supervision.

\vspace{8mm}
\section{Experiments}

\FloatBarrier
\begin{table*}[h]
\centering
\small

\setlength{\tabcolsep}{6pt}
\renewcommand{\arraystretch}{1.15}
    \resizebox{1.0\linewidth}{!}{
    \begin{tabular}{l|cccccc}
    \toprule
    \textbf{Name} & \textbf{ToolAlpaca} & \textbf{SealTool} & \textbf{NexusRaven} & \textbf{API-Bank}  & \textbf{SNIPS} & \textbf{Avg} \\
    \midrule
    \rowcolor{gray!10}\multicolumn{7}{c}{\textbf{Qwen Models}} \\
    \midrule

    \textbf{Qwen2.5-0.5B-Instruct} &  20.17 & 37.07 & 4.71 & 13.85 & 1.57 & 15.47 \\
    \rowcolor{rqBlueBg!50}w/ \toolrzero & 31.58 & 63.95 & 17.61 & 28.00 & 14.29 & 30.57 \\
    \rowcolor{rqBlueBg!50}$\Delta$ & {\color{green!50!black}+11.41} & {\color{green!50!black}+26.88} & {\color{green!50!black}+12.90} & {\color{green!50!black}+14.15} & {\color{green!50!black}+12.72} & {\color{green!50!black}+15.62} \\ 
    & $\uparrow$ 56.57\% & $\uparrow$ 72.51\% & $\uparrow$ 273.89\% & $\uparrow$ 102.17\% & $\uparrow$ 810.19\% & $\uparrow$ 101.03\% \\
    \midrule
    \textbf{Qwen2.5-1.5B-Instruct} & 35.96 & 47.27 & 17.61 & 19.13 & 4.29 & 24.85 \\
    \rowcolor{rqBlueBg!50}w/ \toolrzero & 47.36 & 83.00 & 34.59 & 50.62 & 20.86 & 47.84 \\
    \rowcolor{rqBlueBg!50}$\Delta$ & {\color{green!50!black}+11.40} & {\color{green!50!black}+35.73} & {\color{green!50!black}+16.98} & {\color{green!50!black}+31.49} & {\color{green!50!black}+16.57} & {\color{green!50!black}+22.99} \\ 
    & $\uparrow$ 31.70\% & $\uparrow$ 75.59\% & $\uparrow$ 86.42\% & $\uparrow$ 164.61\% & $\uparrow$ 386.25\% & $\uparrow$ 92.52\% \\
    \midrule
    \textbf{Qwen2.5-3B-Instruct} & 45.61 & 69.72 & 44.33 & 44.95 & 14.28 & 43.97 \\
    \rowcolor{rqBlueBg!50}w/ \toolrzero & 53.51 & 78.23 & 47.8 & 47.94 & 15.57 & 48.50  \\
    \rowcolor{rqBlueBg!50}$\Delta$ & {\color{green!50!black}+7.90} & {\color{green!50!black}+8.51} & {\color{green!50!black}+3.47} & {\color{green!50!black}+2.99} & {\color{green!50!black}+1.29} & {\color{green!50!black}+4.53} \\ 
    & $\uparrow$ 17.32\% & $\uparrow$ 12.21\% & $\uparrow$ 7.83\% & $\uparrow$ 6.65\% & $\uparrow$ 9.03\% & $\uparrow$ 10.30\% \\
    \midrule
    \rowcolor{gray!10}\multicolumn{7}{c}{\textbf{Llama Models}} \\
    \midrule
    \textbf{Llama-3.2-3B-Instruct} & 35.96 & 68.70 & 45.60 & 27.08 & 12.29 & 36.12 \\
    \rowcolor{rqBlueBg!50}w/ \toolrzero & 43.86 & 77.21 & 46.86 & 30.24 & 14.42 & 40.47  \\
    \rowcolor{rqBlueBg!50}$\Delta$ & {\color{green!50!black}+7.90} & {\color{green!50!black}+8.51} & {\color{green!50!black}+1.26} & {\color{green!50!black}+3.16} & {\color{green!50!black}+2.13} & {\color{green!50!black}+4.35} \\ 
    & $\uparrow$ 21.97\% & $\uparrow$ 12.39\% & $\uparrow$ 2.76\% & $\uparrow$ 11.67\% & $\uparrow$ 17.33\% & $\uparrow$ 12.04\% \\
    \bottomrule
    \end{tabular}
    }
    \vspace{-3mm}
    \caption{\textbf{Main results of \toolrzero.} We evaluate \toolrzero\ on five different tool-calling benchmarks, each targeting different aspects of tool use. \hlbox{Highlighted rows} indicate models trained with \toolrzero. {\color{green!50!black}+$\Delta$} reports absolute accuracy improvements over the corresponding instruction-tuned baseline, while percentages denote relative improvement.}
    \label{tab:ttrl-main}
\end{table*}

\FloatBarrier

\vspace{-4mm}

\paragraph{Models.}
We use Qwen-2.5-1.5B-Instruct~\citep{yang2025qwen3} as our primary model, with additional experiments on the 0.5B and 3B variants to analyze scaling within a single architectural family. 
To evaluate cross-family generalization, we also include Llama-3.2-3B-Instruct~\citep{grattafiori2024llama}, which enables us to test our framework across different model scales and foundational architectures: Qwen vs. Llama.

\vspace{-4mm}

\paragraph{Training Details.} 
Both \Generator\ and \Solver\ are optimized with GRPO~\citep{shao2024deepseekmath, guo2025deepseekr1}, initialized from the same base LLM but trained independently.
For our base experiments, we run self-play for three iterations each with 50 steps.
In each iteration, the \Generator\ is trained on 2,000 self-generated samples, then frozen to synthesize 10,000 candidate tasks. 
These are filtered down to 2,000 samples through structural verification, deduplication, and curriculum selection before training the \Solver\ (\Cref{subsec:dataset-construction}).
For the \Generator\ validity reward (\cref{eq:2}) we use $(\lambda_{\text{menu}}, \lambda_{\text{gold}}, \lambda_{\text{value}}){=}(0.4, 0.4, 0.2)$ and for the curriculum reward (\cref{eq:r_diff}), we estimate solver difficulty with $K{=}8$ Monte Carlo samples, using difficulty band $[P_{\text{low}}, P_{\text{high}}]{=}[0.25, 0.75]$ and Gaussian width $\sigma{=}0.12$ as defined in \Cref{fig:rdiff}.
For \Solver\ accuracy reward, we use $(\lambda_{\text{tag}}, \lambda_{\text{parse}}, \lambda_{\text{norm}}){=}(0.3, 0.3, 0.4)$ and $(\lambda_{\text{name}}, \lambda_{\text{key}}, \lambda_{\text{val}}){=}(0.2, 0.3, 0.5)$ with extra-call penalty $\alpha{=}0.25$.
Additional implementation details are provided in the Appendix, including task specification (Appendix~\ref{app:task-specification}), \Generator\ training (Appendix~\ref{app:generator-impl}), dataset construction (Appendix~\ref{app:dataset-construction}), and \Solver\ training (Appendix~\ref{app:solver-impl}).

\vspace{-4mm}

\paragraph{Evaluation.}
We evaluate on five benchmarks spanning diverse tool-calling scenarios: Tool-Alpaca~\citep{tang2023toolalpaca}, Seal-Tools~\citep{wu2024sealtool}, NexusRaven~\citep{srinivasan2023nexusraven}, API-Bank~\citep{li2023apibank}, and SNIPS~\citep{coucke2018snips}.
We follow the previous works~\citep{patil2024gorilla} where all benchmarks are evaluated using Abstract Syntax Tree matching metric, which verifies structural correctness of function names, parameters, and values.
We provide additional details on the evaluation setup in Appendix \ref{app:evaluation}.

\subsection{Results}

\researchq{1}{Can \toolrzero\ enable base LLMs to learn complex tool-calling skills through self-play from scratch?}
We show the main results of \toolrzero\ in \Cref{tab:ttrl-main}. We find that self-play alone is sufficient to induce substantial gains in tool-calling across all evaluated benchmarks. For our primary model, Qwen2.5-1.5B-Instruct, we observe an average improvement of $+22.99$ points (92.52\% relative gain). Notably, these gains are not confined to a single task type; improvements span diverse evaluation settings including single-turn API selection (SealTool), multi-step tool composition (ToolAlpaca, NexusRaven), conversational tool use (API-Bank), and user-intent tracking (SNIPS), indicating that the \textit{learned capabilities generalize across distinct task types rather than overfitting to specific tool distributions} encountered during training. 
These results affirm that self-play RL between the \Generator\ and \Solver\ enables weak base LLMs to self-evolve into general-purpose tool-calling agents, acquiring sophisticated TIR behaviors purely from self-generated experience.

\researchq{2}{How does model scale affect \toolrzero’s tool-calling performance?}
While \toolrzero consistently improves all models across scales, its most striking effect is in \emph{narrowing the capability gap between smaller and larger models}.
After training with \toolrzero, the 0.5B model achieves 30.57 average accuracy, surpassing the 1.5B base model; similarly, the 1.5B model reaches 47.84, exceeding the 3B base model. 
This demonstrates that \toolrzero can effectively elicit latent tool-use capabilities even from models as small as 0.5B, which otherwise exhibit limited tool-calling performance.
We also note that absolute gains are more pronounced for smaller models and we hypothesize that this is because smaller models converge toward their performance upper bound more rapidly than larger models during self-play, which we discuss further in \textbf{Research Question 8}.

\researchq{3}{Is \toolrzero\ robust across different base model families (Qwen vs. Llama)?}
We compare \toolrzero\ across different model types under same scale: Qwen2.5-3B-Instruct and Llama-3.2-3B-Instruct (\Cref{tab:ttrl-main}). They both benefit from our self-play framework. Interestingly, Qwen starts from a stronger baseline and gains a consistent boost (+4.53; $\uparrow$10.30\%), while Llama begins lower yet achieves a comparable post-training level (+4.35; $\uparrow$12.04\%). This shows that \toolrzero\ is model-agnostic generalizes across model families, with gains driven by initial model capability rather than architectural choice.

\begin{takeaway}
    \toolrzero\ enables base LLMs to self-evolve into general-purpose tool-calling agents entirely from scratch without any human data, yielding consistent gains across different benchmarks, model scales, and architectures.
\end{takeaway}

\begin{table*}[th!]
\centering
\renewcommand{\arraystretch}{1.15}
    \resizebox{1.0\linewidth}{!}{
    \begin{tabular}{l c | c c c c c | c}
    \toprule
    \textbf{Model} & \textbf{\#data} &
    \textbf{ToolAlpaca} & \textbf{SealTool} & \textbf{Nexus-Raven} & \textbf{API-Bank} & \textbf{SNIPS} & \textbf{Avg} \\
    \midrule
    \rowcolor{gray!20}
    \multicolumn{8}{c}{\textbf{Agents Trained on Existing Curated Tool-Calling Data}} \\
    \midrule
    Qwen2.5-1.5B-Instruct                & --    & 35.96 & 47.27 & 17.61 & 10.13 & 4.29 & 24.85 \\
    \hspace{1em}$\vdash$w/ xLAM Dataset\small{~\citep{zhang2025xlam}}           & 60k  & 51.75 & 69.05 & 38.68 & 34.65 & \textbf{23.85} & 43.60 \\
    \hspace{1em}$\vdash$w/ Hammer Dataset\small{~\citep{lin2025hammer}}         & 210k & 45.61 & 68.70 & \textbf{51.88} & 33.10 & 19.42 & 43.74 \\
    \hspace{1em}$\vdash$w/ ToolAce Dataset\small{~\citep{liu2025toolace}}       & 12k  & 45.61 & 67.01 & \underline{43.08} & \underline{53.71} & 14.14 & 44.71  \\
    \hspace{1em}$\vdash$w/ ToolRL Dataset\small{~\citep{qian2025toolrl}}        & 4k   & \underline{46.49} & \underline{72.78} & 34.14 & \textbf{62.04} & 14.86 & \underline{46.06} \\
    \midrule
    
    \rowcolor{rqBlueBg}
    \multicolumn{8}{c}{\textbf{\toolrzero Training w/ No Curated Data (Ours)}} \\
    \midrule
    \toolrzero\ (Ours) & \textcolor{blue}{\textbf{0}} &
    \textbf{47.36}$^{\textcolor{green!60!black}{+11.40}}$ &
    \textbf{83.00}$^{\textcolor{green!60!black}{+35.73}}$ &
    34.59$^{\textcolor{green!60!black}{+16.98}}$ &
    50.62$^{\textcolor{green!60!black}{+31.49}}$ &
    \underline{20.86}$^{\textcolor{green!60!black}{+16.57}}$ &
    \textbf{47.84}$^{\textcolor{green!60!black}{+22.99}}$ \\
    \bottomrule
    \end{tabular}
    }
    \begin{minipage}{\linewidth}
\raggedright
\footnotesize\itshape \tiny
\textbf{Note:} For fair comparison, all models are trained using Qwen2.5-1.5B-Instruct with identical hyperparameters on their respective public datasets; see Appendix \ref{app:baselines}.
\end{minipage}
    \vspace{-7mm}
    \caption{\textbf{Performance Comparison with Supervised Baselines.}
    Performance of various models is evaluated on five agentic benchmarks (ToolAlpaca, SealTool, NexusRaven, API-Bank, SNIPS). We use {\color{green!60!black}$+$}
    for absolute accuracy increase from base model.}

    \label{tab:toolzero-others}
\end{table*}

\researchq{4}{How does \toolrzero\ compare to other supervised models trained with human expert data?}
\Cref{tab:toolzero-others} compares \toolrzero\ against models supervised fine-tuned on existing curated tool-calling datasets, where all models share the same Qwen2.5-1.5B-Instruct backbone (see Appendix \ref{app:baselines} for details).
\toolrzero\ achieves 47.84\% average accuracy with \textit{zero} curated data, outperforming methods trained on 4k-210k human-annotated examples. The strongest baseline, ToolRL, excels on specific benchmarks but averages only 46.06\%; suggesting overfitting to distributional patterns in curated datasets. On the other hand, \toolrzero's curriculum adaptively targets the model's evolving weaknesses rather than fixed human priors, avoiding catastrophic forgetting from static data distributions. 
Overall, these results highlight the surprising effectiveness of our approach that pretrained LLMs can generate automated curricula superior to human-designed ones by directly addressing capability gaps the model itself identifies, showing that \textit{the model itself knows best what data it needs}.

\begin{wrapfigure}{r}{0.40\linewidth}
\centering
\vspace{-5mm}
\includegraphics[width=\linewidth]{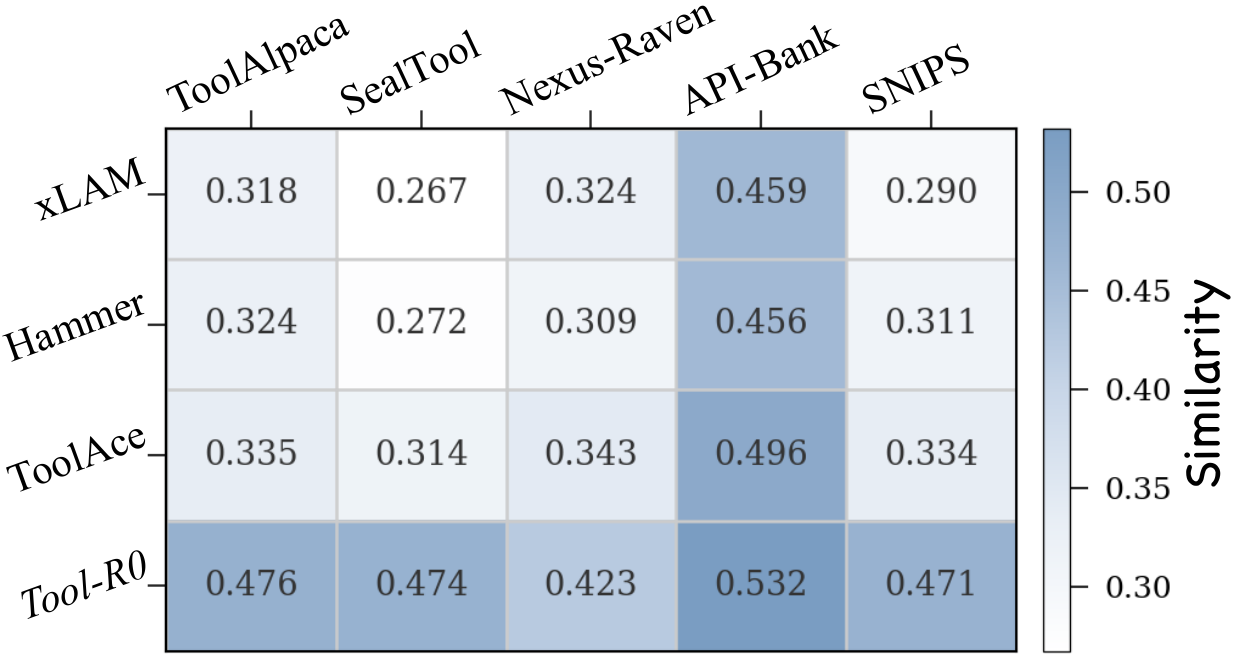}
\vspace{-8mm}
\caption{Self-play coverage analysis.}
\vspace{-5mm}
\label{fig:selfplay-coverage}
\end{wrapfigure}
\textit{4.1 Why does self-play outperform curated supervision?}
To investigate more, we compute pairwise cosine similarity between each training corpus and the test benchmarks (\cref{fig:selfplay-coverage}).
It clearly shows that curated datasets are inherently biased and bounded by their supervision, leading to static training distributions that fail to reflect the evolving needs of agents during training.
In contrast, \toolrzero's self-generated curriculum achieves both the highest average similarity and the most uniform coverage across benchmarks, without any exposure to test data.
This suggests that targeted self-play produces a broader and more balanced coverage of the tool-use distribution, mitigating distribution shift and enabling generalization beyond the limitations of fixed supervision.

\begin{takeaway}
Models trained on adaptive self-generated curricula outperform those trained on human expert data, as self-play produces broader and more balanced training distributions that dynamically target the model's evolving capability gaps rather than relying on static human-designed datasets.
\end{takeaway}

\subsection{Ablation Studies and Analysis}
\label{subsec:analysis}

\begin{table}[h!]
\centering
\small
\resizebox{1.0\linewidth}{!}{
\begin{tabular}{l c cc}
\toprule
\multirow{2}{*}{\textbf{Method}} &
\multirow{2}{*}{\textbf{Average Accuracy}} &
\multicolumn{2}{c}{\textbf{$\Delta$ Accuracy Drop}} \\
\cmidrule(lr){3-4}
 & & \textbf{pp} & \textbf{Relative (\%)} \\
\midrule
\toolrzero\ & 47.84 & -- & -- \\
\midrule
\textit{Ablations} & & & \\
\hspace{1em}$\vdash$ shared weights      & 30.42 & {\color{red!70!black}-17.42} & {\color{red!70!black}$\downarrow$36.41\%} \\
\hspace{1em}$\vdash$ frozen \Generator\ (\textcolor{cyan}{\tiny{\faSnowflake}}) & 41.65 & {\color{red!70!black}-6.19} & {\color{red!70!black}$\downarrow$12.94\%} \\
\hspace{1em}$\vdash$ w/o Difficulty Reward (\Cref{eq:r_diff})     & 43.54 & {\color{red!70!black}-4.30} & {\color{red!70!black}$\downarrow$8.99\%} \\
\hspace{1em}$\vdash$ w/o Gaussian falloff (i.e, with harsh cliffs) (\Cref{eq:r_diff})     & 44.10 & {\color{red!70!black}-3.74} & {\color{red!70!black}$\downarrow$7.82\%} \\
\bottomrule
\end{tabular}
}
\vspace{-3mm}
\caption{\textbf{Ablation analysis relative to \toolrzero.} Absolute changes are reported in percentage points (pp), and relative changes as percentage drop ($\downarrow$\%).}
\label{tab:rzero_ablations}
\end{table}

\researchq{5}{What is the effect of shared vs. separate parameters in \Generator–\Solver\ co-evolution?}
\Cref{tab:rzero_ablations} shows that, despite all variants still improving over the base model, sharing parameters causes a substantial performance drop (-17.42 pp), which we attribute to two complementary factors. 
First, while symmetric role-play with shared weights is effective in closed, single-objective domains such as Go or Poker, tool calling operates over an open-ended, high-entropy action space induced by diverse real-world user requests, requiring inherently asymmetric roles: the \Generator\ must explore and structure an effectively unbounded task distribution, whereas the \Solver\ must reliably execute precise actions under fixed API semantics.
Second, heterogeneous \Generator\ and \Solver\ rewards induce gradient interference under shared parameters: exploration-driven gradients from the \Generator\ conflict with execution-driven gradients from the \Solver, producing unstable representations that neither role can retain, ultimately manifesting as catastrophic forgetting during co-evolution.
Together, these factors suggest that for real-world agentic tasks lacking game-theoretic symmetry, parameter separation is not merely beneficial but essential to prevent co-evolution from collapsing. 

\begin{takeaway}
Parameter separation between \Generator\ and \Solver\ is essential for stable co-evolution in open-ended agentic tasks that operate over high-entropy action spaces, especially when the two roles are optimized with fundamentally different reward objectives.
\end{takeaway}

\researchq{6}{Does the \Generator\ meaningfully improve through self-play, and is its learning essential for effective tool-use performance?}
A core claim of our approach is that the \Generator\ does not merely produce static training data but actively learns to synthesize progressively challenging curricula. We try to validate this claim from three complementary angles: ablation performance, training dynamics, and qualitative illustration.

\textit{6.1 Performance Accuracy.}
To isolate the contribution of \Generator\ learning, we conduct an ablation in which the \Generator\ is frozen after initialization and used only to produce tasks via prompting, while the \Solver\ continues to train.
In this setting, the \Generator\ no longer receives optimization signals from the rewards described in \Cref{subsec:method-generator}.
As shown in \Cref{tab:rzero_ablations}, freezing the \Generator\ leads to a consistent drop of $-6.19$ points in average accuracy of \Solver.
This degradation indicates that performance gains in \toolrzero\ are not solely driven by additional training iterations or static data generation, but critically depend on the \Generator’s ability to generate increasingly \emph{targeted and informative challenges that align with the \Solver’s actual learning needs during self-play}.

\begin{figure}[!t]
    \centering 
    \includegraphics[width=\linewidth]{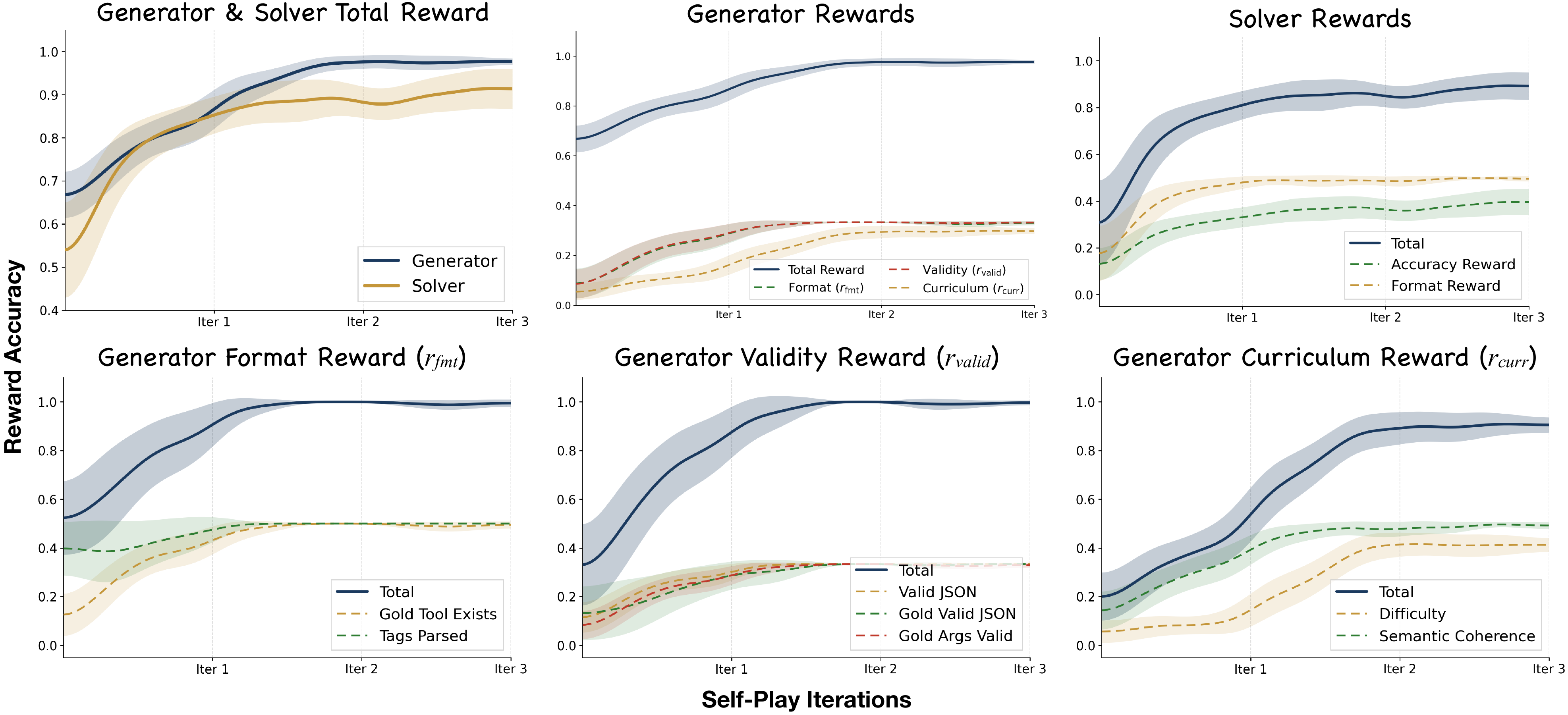}
    \vspace{-5mm}
    \caption{\textbf{RLVR training dynamics of \Generator–\Solver. }\textbf{Top Row:} total reward trajectories for \Generator\ and \Solver\ (left), \Generator\ total reward with format ($r_{\text{fmt}}$), validity ($r_{\text{valid}}$), and curriculum ($r_{\text{curr}}$) components (middle), and \Solver\ total reward with format and accuracy components (right). \textbf{Bottom Row:} detailed \Generator\ reward components, showing format reward (left; $r_{\text{fmt}}$), validity reward (middle; $r_{\text{valid}}$), and curriculum reward decomposed into difficulty and semantic coherence (right; $r_{\text{curr}}$), over self-play iterations.}
    \vspace{-8mm}
    \label{fig:rewards}
\end{figure}

\textit{6.2 Training Signals.}
\Cref{fig:rewards} (Bottom Right) tracks the \Generator's curriculum reward decomposition across training. The difficulty component rises steeply from 0.1 to 0.4 between Iterations 1–2 before plateauing, indicating the \Generator\ learns to produce harder tasks until reaching the \Solver's capacity ceiling. Crucially, semantic coherence remains stable at ~0.5 throughout, confirming that increased difficulty does not sacrifice task validity. The total curriculum reward converges near 0.9, reflecting successful joint optimization of both objectives.

\textit{6.3 Qualitative Evolution.}
\Cref{fig:gen-evolution-travel-easy} and \Cref{fig:gen-evolution-travel-hard} contrast \Generator\ outputs from early versus late trainings stages. At first iteration of self-play, the \Generator\ produces minimal tasks: single-sentence requests, one available tool with two parameters, and a single-call solution. 
By Iteration 3, complexity increases across all dimensions—user requests contain five explicit constraints (dates, passenger count, cabin class, hotel location), the tool menu expands to two functions with eleven total parameters, and gold solutions require two coordinated tool calls with cross-task dependencies (flight arrival date must precede hotel check-in). 
This progression from surface-level to compositional multi-step planning demonstrates learned curriculum generation rather than random variation.

\researchq{7}{What is the specific contribution of difficulty reward in \Generator\ self-play?}
Our curriculum reward $r_{\text{curr}}$ is designed to steer the \Generator\ toward producing tasks near the \Solver's competence frontier via the band-pass difficulty signal in \Cref{eq:r_diff}. To test whether this calibration matters, we ablate $r_{\text{diff}}$ entirely. As shown in \Cref{fig:rdiff} (last row), removing this reward decreases average accuracy by $4.30$ pp ($\downarrow$8.99\% relative), demonstrating that solvable task generation alone is insufficient without calibrating difficulty.
The training dynamics in \Cref{fig:rewards} (bottom right) corroborate this finding: both difficulty and semantic coherence components rise steadily across iterations, indicating that the \Generator\ learns to occupy the target difficulty band rather than collapsing to degenerate modes. Together, these results show that difficulty-aware curriculum shaping is a core mechanism enabling reliable self-play, allowing the \Generator\ to produce targeted challenges that meaningfully advance the \Solver’s capabilities.

\textit{7.1 Role of smooth Gaussian transitions in difficulty reward.} Our band-pass reward in \Cref{eq:r_diff} uses smooth Gaussian transitions outside our band $[p_{\text{low}}, p_{\text{high}}]$ rather than hard cutoffs. We hypothesize that this design is critical for stable training: when the \Generator\ produces a task slightly outside the target band, the smooth transition still provides a signal proportional to how far it has drifted and can guide it back toward the desired difficulty range. We ablate this by replacing the Gaussian falloffs with a rectangular filter that assigns reward $1$ inside the band and $0$ outside. As shown in \Cref{tab:rzero_ablations} (last row), it degrades accuracy, showing that harsh reward clips destabilize learning by eliminating informative feedback near the competence frontier, while smooth transitions enable more stable learning.

\begin{takeaway}
Effective self-play requires an actively learning \Generator\ guided by difficulty-aware curriculum and smooth reward signals, as static data generation or uncalibrated task difficulty fails to produce the targeted challenges necessary for sustained tool-use improvement.
\end{takeaway}

\vspace{2mm}

\begin{wrapfigure}{r}{0.46\linewidth}
\centering
\vspace{-5mm}
\includegraphics[width=\linewidth]{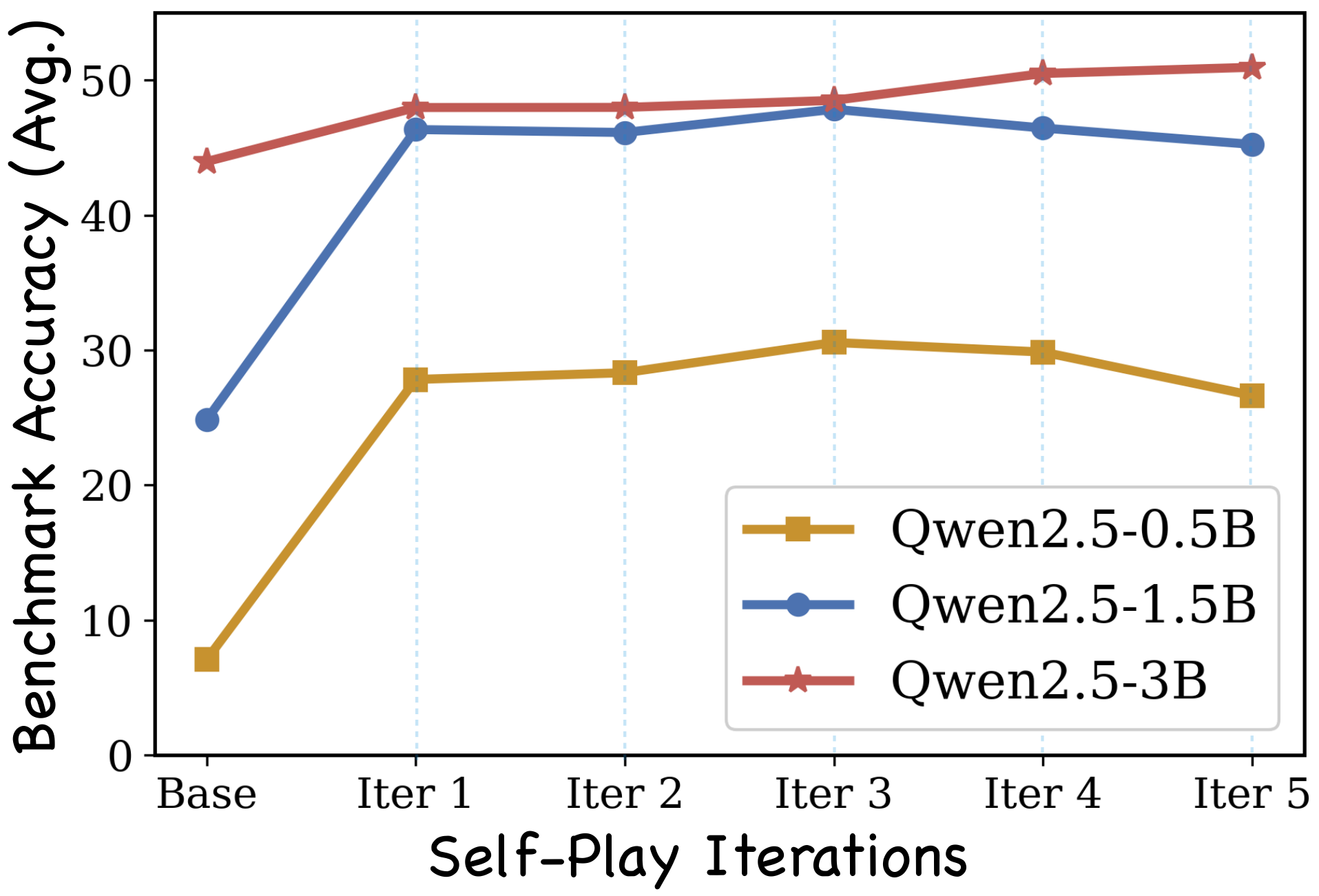}
\vspace{-7mm}
\caption{Self-play convergence for extended iterations across model scales.}
\vspace{-3mm}
\label{fig:selfplay-convergence}
\end{wrapfigure}
\researchq{8}{When does self-play saturate, and what factors limit continued improvement?}
To understand self-play convergence, we extend training to five iterations and analyze performance trends across model scales. 
We observe rapid gains after the first iteration, with accuracy typically peaking around the third for smaller models, after which improvements saturate or even slightly degrade.
The rapid stabilization in low-capacity models is consistent with early convergence to a Nash-like equilibrium and a potential knowledge boundary, where \Generator\ and \Solver\ become mutually aligned. 
In contrast, the 3B model exhibits a more steady and continuous improvement with no signs of saturation, suggesting that higher model capacity delays convergence and that the model has potential to accumulate further gains from additional self-play iterations.
This explains the pattern observed in \textbf{Research Question 2}: smaller models converge toward their upper bound more quickly, yielding large initial gains but limited headroom for further improvement, while higher-capacity models evolve more gradually but consistently under self-play.

\vspace{2mm}

\researchq{9}{Can \toolrzero serve as an effective mid-training strategy to amplify post-training?}
Recent works suggests that mid-training can incentivize RL scaling in ways invisible from base model evaluations for abstract reasoning~\citep{wang2025octothinker}.
\begin{wrapfigure}{r}{0.38\linewidth}
\centering
\vspace{-4.5mm}
\includegraphics[width=\linewidth]{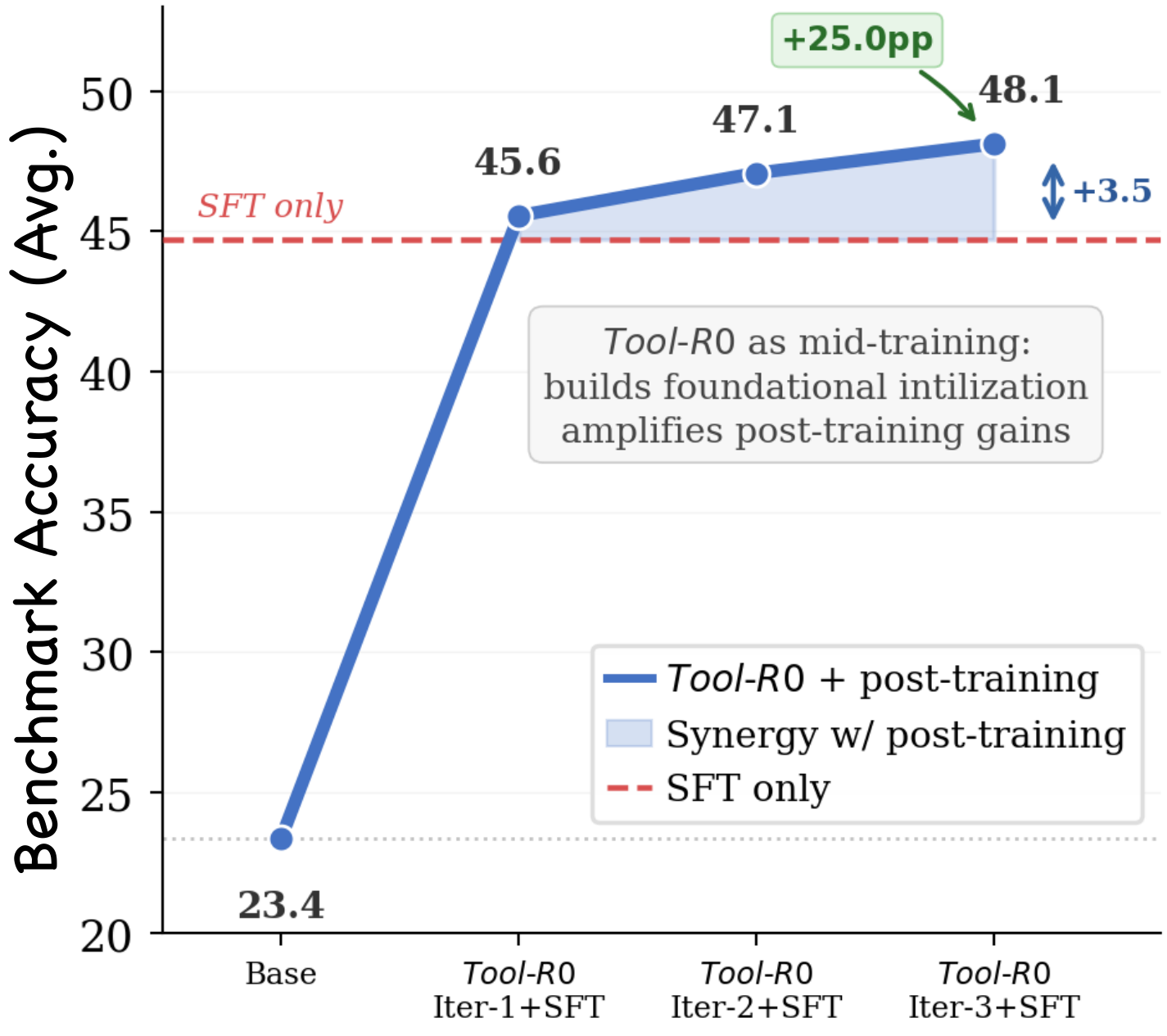}
\vspace{-8mm}
\caption{\toolrzero as Mid-training.}
\vspace{-4.5mm}
\label{fig:selfplay-sft}
\end{wrapfigure}
We investigate whether a similar paradigm holds for tool-calling tasks: can iterative self-play serve as continued pre-training that strengthens later supervised post-training on human data? To test this, we treat each \toolrzero iteration as a mid-training checkpoint and fine-tune it with ToolACE~\citep{liu2025toolace}, a well-established tool-use dataset which is also used in our baselines. As shown in \cref{fig:selfplay-sft}, \toolrzero followed by supervised post-training surpasses the SFT baseline from the first iteration and continues to scale with each subsequent round. After iteration three, it outperforms both post-training alone and standalone \toolrzero, showing that self-play as continued pre-training incentivizes a stronger foundation where supervised post-training extracts more from the same data. This suggests that self-play can serve as a scalable initial stage to strengthen supervised alignment later.

\vspace{2mm}

\begin{takeaway}
\toolrzero\ serves as an effective mid-training stage that strengthens the foundation for supervised post-training, enabling the model to better explore and extract from the same human-curated data, outperforming human supervision alone.
\end{takeaway}

\researchq{10}{What do the reward dynamics reveal about the learning behavior and co-evolutionary stability of \Generator–\Solver\ self-play?}
We examine reward trajectories across self-play iterations to understand the learning dynamics of \Generator–\Solver\ co-evolution (\Cref{fig:rewards}). Several patterns emerge: first, the \Generator\ converges faster than the Solver, and \Generator\ total reward reaches $\sim$0.98 by iteration two, while \Solver\ reward stabilizes around 0.90. This asymmetry reflects the inherent difficulty gap-synthesizing valid tasks is easier than solving them. 
Second, reward components exhibit a clear learning hierarchy. 
For the \Generator, format compliance saturates within the first iteration, validity rewards improve steadily through iteration two, and curriculum rewards show the steepest growth trajectory. 
This ordering suggests that \Generator\ first learns structural constraints, then internal consistency, and finally calibrating task difficulty. 
For the \Solver, format rewards rise faster than accuracy rewards, with accuracy remaining the performance bottleneck even at convergence. 
Third, the curriculum reward decomposition reveals stable co-evolution: difficulty increases sharply from 0.1 to 0.5 between iterations one and iteration two as the \Generator\ learns to challenge the improving \Solver, yet semantic coherence rises gradually rather than collapsing. 
This confirms that the \Generator\ produces progressively harder tasks without sacrificing validity. The convergence of both agents toward high total reward with narrowing gap suggests that they both approach a stable equilibrium where \Generator\ output matches \Solver\ capacity---consistent with the saturation behavior observed in downstream evaluation.

\begin{wrapfigure}{r}{0.40\linewidth}
\centering
\vspace{-5mm}
\includegraphics[width=\linewidth]{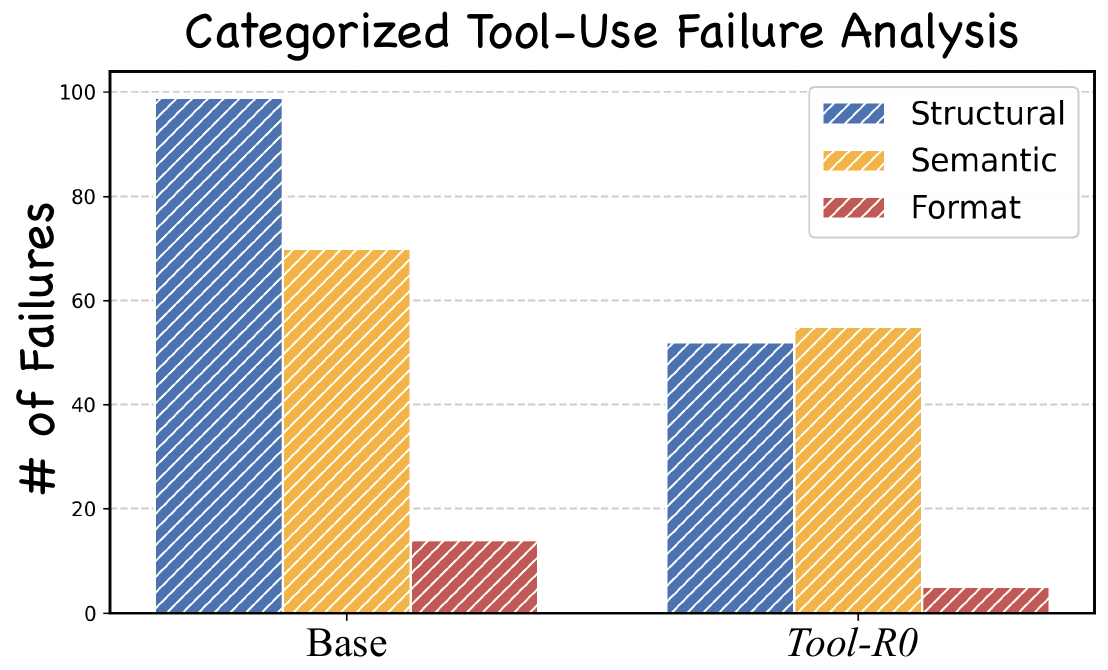}
\vspace{-8mm}
\caption{Categorical analysis of tool failures for base model and \toolrzero.}
\vspace{-3mm}
\label{fig:selfplay-errors}
\end{wrapfigure}
\researchq{11}{What are the common tool-use failures of the base model and how does \toolrzero\ address them?}
To understand what self-play improves at a fine-grained level, we group failures into three categories: \emph{structural errors} (wrong tool name, incorrect number of calls, extra or missing arguments), \emph{semantic errors} (wrong argument values, missing gold keys), and \emph{format errors} (malformed or unparseable JSON). 
As shown in \cref{fig:selfplay-errors}, the base model fails predominantly due to structural errors.
These represent the most critical failure modes: selecting the wrong tool indicates a fundamental misunderstanding of tool capabilities, producing incorrect number of calls shows an inability to decompose requests into appropriate multi-tool plans, and hallucinating extra arguments reflects poor adherence to tool schemas. 
\toolrzero\ reduces these by nearly half, confirming that iterative self-play builds stronger TIR capabilities that transfer across tool selection, multi-step planning, and schema compliance. 
Semantic errors also decrease but remain the dominant bottleneck while format errors where the base model is already relatively strong are reduced to near elimination.

\section{Discussion}
\label{sec:discussion}

\subsection{Conclusion}
\vspace{-1mm}
We present \toolrzero, a self-play RL framework that enables base LLMs to self-evolve into general-purpose tool-calling agents with zero human data. 
Across diverse benchmarks and model scales, we show that self-play supports consistent self-improvement and can match or surpass supervised baselines trained on static human data. 
Our analysis reveals core findings that govern self-play for agentic learning: i) self-play alone suffices to incentivize complex agentic skills from weak priors, (ii) self-generated curricula produce broader training distributions than static human supervision, (iii) role separation is essential for stable co-evolution in high-entropy action spaces, and (iv) difficulty-aware reward shaping is critical to sustain learning across iterations.
While preliminary, these results suggest that \toolrzero can successfully convert weak base LLMs to general-purpose tool-calling agents that can self-evolve across different domains without requiring any external data.
This points toward a future in which agents continuously acquire new capabilities across tools and environments without human intervention, bringing us one step closer to the vision of artificial superintelligence.

\subsection{Challenges}
\vspace{-1mm}
While \toolrzero\ demonstrates promising results for training tool-using LLM agents through zero-data self-play towards superintelligent agents that can continuously improve themselves; it still remains an early step and several limitations warrant discussion:
\begin{itemize}[topsep=-1.5pt, leftmargin=10pt, itemsep=-0pt]

\item \textbf{Model Scale \& Reward Robustness.}  
Smaller models sometimes exhibit imperfect instruction and policy following, which can lead to rare reward-hacking behaviors that pass verifiable checks but yield low-quality supervision for the \Solver. We mitigate this with semantic coherence checks and validity rewards, though these signals may miss subtle failures. In practice, such cases are infrequent and diminish with stronger base models, though larger-scale settings may require more careful reward calibration and prompt design.

\item \textbf{Early Self-Play Saturation.}  
For lower-capacity models, self-play often converges within a few iterations, suggesting an early alignment between \Generator\ and \Solver\ near a competence boundary. In contrast, higher-capacity models show slower but more sustained improvement, indicating that saturation behavior depends on model scale and intrinsic reasoning capacity.We believe that a deeper investigation into the equilibrium dynamics and convergence properties of the \Generator\ and \Solver, particularly as an interpretable mechanism underlying self-play, is a promising direction for future work.

\item \textbf{Curriculum Signal Efficiency.}  
The curriculum reward estimates task difficulty by querying the \Solver\ multiple times, which increases computational cost and may imperfectly correlate with actual learnability. While our verifiers prevent unsatisfiable tasks in practice, alternative signals more directly tied to learning progress—such as loss dynamics or gradient-based measures could further improve curriculum alignment~\citep{koh2017understanding}.

On the other hand, given our computational and hardware constraints, this preprint reports results without comprehensive multi-run standard error analysis. Preliminary repetitions show consistent trends with low variance across runs, and we will include detailed statistical analysis in the final version.
\end{itemize}

\subsection{What is Next}
\vspace{-1mm}
Beyond the limitations discussed above, our results suggest several promising directions for pushing self-play toward more capable, general-purpose tool-using agents and, ultimately, end-to-end superintelligent systems:
\begin{itemize}[topsep=-1.5pt, leftmargin=10pt, itemsep=-0pt]
    \item \textbf{Richer difficulty feedback beyond solver consistency.}
    Our curriculum signal estimates difficulty via the \Solver's stochastic consistency, which can be noisy and sparse for guiding \Generator\ learning.
    A promising direction is to use more informative and interpretable feedback, such as loss-based or gradient-based signals, and semantic error attribution (e.g., \emph{what} failed and \emph{why}).
    Plateau-shaped rewards over loss (or calibrated margin objectives) could provide smoother learning signals than binary success, while still preserving verifiability.

    \item \textbf{Breaking the knowledge boundary with external signals.}
    In our current setup, \Generator\ and \Solver\ co-evolve from the same base model, relying on latent knowledge to bootstrap curricula.
    While effective for rapid domain adaptation, self-play often saturates near a knowledge boundary where both roles reach a Nash-like equilibrium.
    Future work could introduce a stronger third-party teacher (e.g., a higher-capability LLM or tool-backed oracle) that diagnoses persistent failure modes and injects targeted missing knowledge when progress stalls.

    \item \textbf{Grounding as a Prerequisite for Diverse Generation.} We consistently observed that explicit grounding (such as target domain, tool count, or interaction format) is essential to avoid mode collapse and maintain generation quality. Without such constraints, generators produce repetitive, narrowly distributed samples. This aligns with prior findings on structured constraints in synthetic data generation~\citep{liu2025spice}. Future self-play methods could systematically investigate environmental grounding as a mechanism for maintaining diversity while providing reliable learning signals. Understanding when and how to introduce such constraints may be key to scaling self-play beyond narrow benchmarks.

    \item \textbf{Quantitative metrics for generation quality.}
    A key bottleneck in data-generation-based training is the lack of reliable, automatic measures of sample quality beyond qualitative inspection or downstream accuracy.
    Developing quantitative metrics for task realism, ambiguity, coverage, and label reliability would enable better filtering, more stable curricula, and direct reward shaping for the \Generator.

\end{itemize}

\bibliography{colm2026_conference}
\bibliographystyle{colm2026_conference}

\newpage
\appendix
\section*{Appendix}
\label{app:main}
\section{Related Work Extended}
\label{app: related-work-extened}

\paragraph{Tool Learning with LLMs.}
Tool-integrated reasoning (TIR) enables LLMs to ground parametric knowledge through external tools such as APIs, databases, and software functions~\citep{qu2025toolsurvey}.
Early work primarily focused on benchmarking and evaluation, measuring tool selection, argument generation, and execution correctness across curated datasets~\citep{li2023apibank, tang2023toolalpaca, srinivasan2023nexusraven, wu2024sealtool, patil2024gorilla}.
Building on these benchmarks, subsequent efforts emphasized data construction and supervised fine-tuning, producing large-scale, human- or model-generated instruction corpora to teach tool use~\citep{mitra2024agentinstruct, chen2024agentflan, zhang2024agentohana}. More recent systems combine such data with stronger post-training pipelines to improve robustness and generalization~\citep{zhang2025xlam, liu2025toolace, lin2025hammer, acikgoz2025coalm}, while reinforcement learning has been applied as an auxiliary stage to refine function-calling accuracy~\citep{qian2025toolrl}.

Despite steady progress, all existing approaches fundamentally rely on curated supervision---either explicit demonstrations, synthetic instruction data, or static task distributions. This reliance limits scalability to new domains where realistic user requests, correct tool traces, and verifiable outcomes are unavailable.
In contrast, we study tool learning under a strict zero-data assumption, where no demonstrations, prompts, or external task corpora are accessible, and show that tool-use skills can emerge purely through self-play and autonomous curriculum generation.

\paragraph{Self-Play in LLMs.}
Self-play has long driven advances in AI for superintelligence, from early curiosity-driven two-agent setups~\citep{schmidhuber2011PowerPlayTA} to game mastery in TD-Gammon~\citep{tesauro1995tdgammon}, AlphaGo~\citep{silver2016alphago, Silver2017alphagozero}, and CICERO~\citep{brown2022cicero}.
In LLMs, self-play initially focused on alignment via methods like SPIN~\citep{chen2024spin} evolving to capability enhancement in verifiable domains such as code generation with Coder-Tester pairs~\citep{wang2025codetestevolve} or adaptive problem creation from zero data~\citep{zhao2025absolute, fang2025serl}.
Recent efforts like SPAG~\citep{cheng2024spag}, SPC~\citep{chen2025spc}, Genius~\citep{xu2025genius}, and SPIRAL~\citep{liu2025spiral} use games or seeded tasks for reasoning gains, yet pure self-play often stalls---R-Zero shows marginal gains and degrades after iterations~\citep{huang2025r-zero}, Absolute Zero enhances performance in coding and mathematics but limited to coding-bound as verifiable environment~\citep{zhao2025absolute}, and Agent0 relies on single-tool Python calls~\citep{xia2025agent0}. 
More importantly, these methods are restricted to general knowledge or math environments and remain largely underexplored for agentic tasks that require complex, real-world tool use. 
While \citet{zhou2025selfchallenging} proposes Code-as-Task generation for targeted and more challenging training data, it is limited by code-based verification (similar to Absoulte Zero~\citep{zhao2025absolute}), depends on structured tools and environments rather than a true zero-data paradigm, and optimizes only the \Generator, precluding genuine \Generator–\Solver\ co-evolution. 
Crucially, none of these approaches address general-purpose tool-using agents operating over heterogeneous, real-world APIs, where actions are high-entropy, verification is execution-based, and task distributions must evolve alongside model competence.
Moreover, several recent works optimize only the task generator as challenger~\citep{zhou2025selfchallenging}, precluding genuine co-evolution and often leading to unstable or saturated learning dynamics.

Our work fills this gap by introducing a dual-agent self-play framework in which a task-generating \Generator\ and an executing \Solver\ co-evolve symbiotically under complementary reward signals. Unlike prior self-play methods, \toolrzero\ targets tool-integrated reasoning across diverse domains, operates in a fully zero-data regime, and demonstrates sustained capability gains driven by adaptive, self-generated curricula rather than static task generation.

\section{Grounded Task Specification}
\label{app:task-specification}

\subsection{Formal Definitions}
To ensure domain-agnostic yet controllable task generation, the \Generator\ is conditioned on an explicit domain configuration specification that constrains both the semantic scope and structural properties of each generated task. See \Cref{fig:domain-config} for the standardized domain configuration templates used in our experimental setup.
This design is conceptually related to the grounding strategy of \citet{liu2025spice}, where constrained generation improves specificity, diversity, and robustness to hallucination.
However, unlike \citet{liu2025spice}, which grounds generation in large, static documents, our approach relies on lightweight, dynamically adjustable configurations that encode user-level preferences.
This enables rapid adaptation across users and domains while preserving control, making grounding a flexible interface rather than a fixed knowledge source. In our setting, each training example is associated with a sampled specification
\begin{equation}
    s = (d, c, m, n)
\end{equation}
where $d$ denotes the task domain, $c$ the interaction context type, $m$ the number of available tools, and $n$ the number of gold tool calls.

\subsection{Domain Configurations}
\begin{figure}[h!]
\centering
\begin{lstlisting}[language=Python,basicstyle=\ttfamily\small]
DOMAIN_WEIGHTS = {
    "finance": 0.03125, "healthcare": 0.03125, 
    "productivity": 0.03125, "retail_ecommerce": 0.03125, 
    "scheduling": 0.03125, "database": 0.03125,
    "cloud_infrastructure": 0.03125, "system": 0.03125, 
    "programming": 0.03125, "geolocation": 0.03125, 
    "logistics": 0.03125, "communication": 0.03125,
    "iot": 0.03125, "cybersecurity": 0.03125, "insurance": 0.03125,
    "legal": 0.03125, "news": 0.03125, "weather": 0.03125,
    "sports": 0.03125, "entertainment": 0.03125, 
    "education": 0.03125, "real_estate": 0.03125, 
    "food_ordering": 0.03125, "translation": 0.03125,
    "utilities": 0.03125, "government": 0.03125, 
    "memory_management": 0.03125, "web_search": 0.03125, 
    "social_media": 0.03125, "math": 0.03125,
    "vehicle_control": 0.03125, "travel": 0.03125
}
\end{lstlisting}
\caption{\textbf{Domain sampling configuration.} 
The Generator samples task domains from a fixed set of functional and agentic categories using a uniform prior.
Weights are treated as unnormalized sampling coefficients rather than probabilities, currently uniform.}
\label{fig:domain-config}
\end{figure}

The \Generator\ receives this specification through a structured system prompt (See \Cref{fig:generator-prompt-template}) that enforces strict adherence to the desired domain and interface. The prompt requires the model to (i) generate a natural user question grounded in the \textit{specified domain}, (ii) define an \textit{explicit tool menu of size} $m$ with JSON-verifiable schemas, and (iii) provide exactly $n$ \textit{number of gold tool calls} with flat primitive arguments.\footnote{To prevent ungrounded or templated outputs, the prompt explicitly forbids meta-instructions, placeholders, or abstract descriptions, and requires all argument values in the gold tool calls to appear verbatim in the user question.} This design allows us to control generation without providing any task-level supervision, while ensuring that every generated instance admits automated verification and execution-based feedback. Specifications $s=(d,c,m,n)$ are sampled independently for each training example. The sampling strategy is intentionally \textit{non-uniform}, reflecting realistic tool-use distributions and emphasizing domains where precise tool selection and argument grounding are critical.

\subsection{Sampling Strategy and Hyperparameters}
\paragraph{Domain sampling.} Domains are drawn from a fixed weighted uniform distribution over more than 30 functional and agentic categories, including finance, healthcare, scheduling, databases, cloud infrastructure, and system utilities as in \Cref{fig:domain-config}.
Crucially, this \textbf{distribution is entirely user-defined}: practitioners specify both the set of domains and their relative weights, allowing the \Generator’s data curriculum to directly reflect task importance, deployment priorities, or safety constraints.
Higher weights may be assigned to precision-critical domains (e.g., finance, healthcare, productivity), while exploratory or open-ended domains (e.g., web search, social media) can be intentionally down-weighted.
We treat these values as unnormalized sampling preferences rather than probabilities, enabling flexible reconfiguration without retraining or architectural changes.

\paragraph{Hyperparameters for Task Specifications.} The interaction context is sampled as: (i) single-turn with probability 0.9 and (i) multi-turn with probability 0.1. Multi-turn prompts embed short conversational histories inside the user question and are restricted to a single gold tool call to reduce ambiguity. For single-turn contexts, the number of gold calls is sampled as $n=1$ with probability 0.8 and $n=2$ with probability 0.2. For multi-turn contexts, $n$ is fixed to 1. When $n>1$, the tool menu size is sampled from $\{3,4,5\}$ to limit combinatorial ambiguity. When $n=1$, the menu size is sampled from two buckets: small menus (2–4 tools) or larger menus (5–8 tools), mimicking evaluation benchmarks with varying tool density. This specification sampling scheme encourages compositional diversity while maintaining solvability under automated verification.
\section{Generator Implementation Details}
\label{app:generator-impl}

This section specifies the hyperparameters and implementation details used to instantiate the \Generator\ training procedure described in \Cref{subsec:method-generator}. \Cref{tab:reward-hyperparams-generator} summarizes all reward-related hyperparameters used in \Generator\ training.
Unless otherwise stated, all values are fixed across experiments.

\subsection{Training Setup}
\Cref{tab:hyperparams-gen-solver} summarizes the \Generator's training hyperparameters. Our \Generator\ is trained with Grouped Relative Policy Optimization (GRPO)~\citep{shao2024deepseekmath, guo2025deepseekr1} using the TRL~\citep{vonwerra2020trl} implementation.
Training is performed with HuggingFace Accelerate and DeepSpeed ZeRO-3~\citep{rajbhandari2020zero} in mixed precision (\texttt{bfloat16}) on three GPUs.
We use a per-device batch size of $2$ with gradient accumulation over $4$ steps, resulting in a global batch size of $24$ sequences per update.
The optimizer is AdamW~\citep{loshchilov2017adamw} with a fixed learning rate of $1\times10^{-6}$.
We train for 50 steps per run and we sample $4$ generations per prompt during GRPO updates to compute relative advantages.
The maximum completion length is set to $4096$ tokens, which accommodates full task specifications including tool menus and gold tool calls.

\begin{table}[h!]
\centering
\begin{tabular}{lcc}
\toprule
\textbf{Hyperparameter} & \Generator\ & \Solver\ \\
\midrule
Dataset Size                                     & 2000 & 2000 \\
Global Batch Size                                & 24  & 32 \\
Optimizer                                        & AdamW  & AdamW \\
Learning Rate                                    & $1 \times 10^{-6}$  & $1 \times 10^{-6}$ \\
Weight Decay                                     & $1 \times 10^{-2}$  & $1 \times 10^{-2}$ \\
KL Penalty Coefficient ($\lambda_{\mathrm{KL}}$) & $1 \times 10^{-2}$  & $1 \times 10^{-2}$ \\
Max Steps                                        & 50  & 50 \\
Number of Group Rollouts                         & 4  & 4 \\
Rollout Temperature                              & 1.0  & 1.0 \\
Precision                                        & BFloat16  & BFloat16 \\
Max Sequence Length                              & 4096  & 4096 \\
\bottomrule
\end{tabular}
\caption{\textbf{Training hyperparameters.} Details of the main training configurations for the \Generator\ and \Solver\ during Self-Play iterations.}
\label{tab:hyperparams-gen-solver}
\end{table}

\subsection{Reward Hyperparameters}

\paragraph{Structured output interface.}
Each \Generator\ completion is required to emit exactly four tagged blocks: \thinkS, \questionS, \menuS, and \answerS, as defined in §\ref{subsec:method-generator}.
The \menuS\ block must parse as a JSON list of tool specifications, and the \answerS\ block must parse as a JSON list of tool calls with flat primitive arguments only.
This constraint enables deterministic parsing and execution-based reward computation.

\paragraph{Reward composition.}
The \Generator\ is trained with three reward components: format reward $r_{\text{fmt}}$ (\Cref{eq:1}), validity reward $r_{\text{valid}}$ (\Cref{eq:2}), and curriculum reward $r_{\text{curr}}$ (\Cref{eq:4}).

The format reward is a sum of three binary indicators as defined in \Cref{eq:1}: tag completeness ($\mathbb{I}_{\text{tags}}$), tool-menu JSON validity ($\mathbb{I}_{\text{tools-json}}$), and gold-call JSON validity ($\mathbb{I}_{\text{gold-json}}$).

The validity reward assigns weights $(\lambda_{\text{menu}}, \lambda_{\text{gold}}, \lambda_{\text{value}}) = (0.4, 0.4, 0.2)$ to three checks: (i) the gold tool name exists in the parsed menu, (ii) all schema-required parameters are present in the gold call, and (iii) every non-trivial argument value (excluding booleans and nulls) appears as a word-boundary match in the generated question ($\operatorname{vals}(a^\star) \hookrightarrow q$), as described in \Cref{eq:2}.
This acts as a compiler-like gate: the first two checks prevent hallucinated calls to non-existent tools or calls with missing arguments, while the third acts as a semantic anchoring constraint that ties the gold answer back to the generated question, discouraging solutions whose arguments have no evidential basis in the task.
When \answerS\ contains a list of calls, we canonicalize all calls into a normalized tool-call representation for verification.

The curriculum reward $r_{\text{curr}} = r_{\text{diff}} + r_{\text{sem}}$ is an unweighted sum of the difficulty and semantic alignment components, as defined in \Cref{eq:4}.

\paragraph{Difficulty estimation.}
As described in \Cref{eq:r_diff}, solver-calibrated difficulty is estimated using Monte Carlo sampling.
We query the current \Solver\ $K=8$ times per task with temperature $0.7$ and maximum generation length $2048$ tokens.
Note that this temperature ($0.7$) is used exclusively for difficulty estimation during \Generator\ reward computation, and is distinct from the \Solver's training rollout temperature ($1.0$) listed in \Cref{tab:hyperparams-gen-solver}.
Tasks with empirical success probability $\hat{p}_{\text{succ}} < 1/K$ (i.e., no solver sample matches the gold tool-call) receive zero difficulty reward, as they likely represent ill-posed, ambiguous, or unsolvable generations that provide no meaningful learning signal.
For solvable tasks, difficulty is shaped using the band-pass function in \Cref{eq:r_diff} with parameters $P_{\text{low}}=0.25$, $P_{\text{high}}=0.75$, and $\sigma=0.12$.

\paragraph{Semantic alignment.}
Semantic alignment between the generated user question and the gold tool call is evaluated using the \Solver\ as a judge.
The \Solver\ assigns an integer score in $\{1,\dots,5\}$ for semantic coherence (\Cref{fig:semantic-alignment-prompt-template}), which is normalized to $[0,1]$ as $r_{\text{sem}}=(s-1)/4$.
This signal penalizes vague or templated questions even when the corresponding tool call is syntactically valid.

\paragraph{Normalization.}
Gold tool calls are normalized into a canonical $(\texttt{name}, \texttt{arguments})$ representation prior to verification.
Schema validation enforces presence of required parameters but does not perform strict type checking to keep reward computation inexpensive.

\begin{table}[t]
\centering
\small
\begin{tabular}{lc}
\toprule
\textbf{Hyperparameter} & \textbf{Value} \\
\midrule
\multicolumn{2}{l}{\textit{Validity Reward} (\Cref{eq:2})} \\
$\lambda_{\text{menu}}$ (tool existence)          & $0.4$ \\
$\lambda_{\text{gold}}$ (required arguments)       & $0.4$ \\
$\lambda_{\text{value}}$ (value grounding)         & $0.2$ \\
\midrule
\multicolumn{2}{l}{\textit{Curriculum Reward} (\Cref{eq:r_diff})} \\
Solver samples ($K$)                               & $8$ \\
Solver temperature (difficulty estimation)         & $0.7$ \\
Solver max tokens (difficulty estimation)          & $2048$ \\
Difficulty band $(P_{\text{low}},P_{\text{high}})$ & $(0.25,\,0.75)$ \\
Gaussian width $\sigma$                            & $0.12$ \\
Semantic alignment scale                           & $\{1,\dots,5\} \to [0,1]$ \\
\bottomrule
\end{tabular}
\caption{\textbf{\Generator\ reward hyperparameters.} Hyperparameter values for the validity and curriculum reward components used during \Generator\ training (\Cref{subsec:method-generator}). The format reward (\Cref{eq:1}) uses unweighted binary indicators and has no tunable parameters.}
\label{tab:reward-hyperparams-generator}
\end{table}
\section{Details of Solver Dataset Construction}
\label{app:dataset-construction}

After training the \Generator, we freeze it and use it purely as a task synthesizer conditioned on the same control specifications described in \Cref{subsec:task-config}. Starting from these domain-conditioned inputs, we construct \Solver\ training data through a three-stage pipeline consisting of \emph{generation and deduplication}, \emph{cross-verification}, and \emph{difficulty probing and selection}.

\subsection{Specification-Grounded Generation and Deduplication}
We first sample a large pool of $10{,}000$ candidate tasks from the frozen \Generator, producing structured triples of user requests, tool menus, and gold tool calls. To avoid training bias caused by repeated patterns, we remove near-duplicate samples via canonicalized signatures derived from question–tool–call combinations, producing a large but non-redundant candidate pool.

\subsection{Solver-Based Cross-Verification}
To further increase the reliability of generated pseudo-labels, we sample each candidate multiple times using the \Solver\ and measure agreement between predicted and generated gold tool calls, retaining only tasks with consistent solutions and discarding instances with low agreement. This procedure follows the principle that reproducible answers provide more reliable supervision, as consistently reproduced solutions are more likely to correspond to correct supervision signals~\citep{huang2023llmsselfimprove, zuo2025ttrl, acikgoz2025tt-si}. Consequently, this stage filters ambiguous or noisy pseudo-labels and retains only tasks whose solutions provide reliable signal.

\subsection{Difficulty Probing and Curriculum Selection}
From the verified pool, we estimate task difficulty via \Solver\ pass@$K$ success rates and group the generated tasks into easy, medium, and hard buckets. Samples are then selected to preserve domain diversity while maintaining a balanced difficulty mix, preventing bias toward trivially solvable tasks. Through this pipeline, the initial $10{,}000$ candidates are filtered down to $2{,}000$ samples that form the final \Solver\ training set for each self-play iteration.

The selected data are organized into a staged curriculum progressing from easier to harder instances at the training batch level: early batches are composed primarily of easy samples, while harder tasks are progressively introduced in later batches.
Data difficulty is critical for effective RL, as training data must align with the model's current capabilities to avoid learning failure~\citep{liu2025understanding, wang2025octothinker}.
When the \Solver\ is exposed to tasks far beyond its current competence too early, policy gradients become noisy and uninformative, leading to unstable optimization or degenerate solutions.
By categorizing synthesized tasks by \Solver\ answer consistency and progressively exposing the \Solver\ to harder problems across batches, we ensure that the model builds foundational tool-calling skills on reliable examples before tackling compositional multi-tool scenarios that require those skills as prerequisites.

This design ensures that the final dataset is simultaneously diverse, semantically valid, and appropriately challenging, while avoiding noisy or degenerate pseudo-labels that could destabilize \Solver\ training. In contrast to static synthetic pipelines, our procedure continuously grounds generation quality in \Solver\ behavior, yielding a self-consistent data distribution suitable for stable agentic learning.
\section{Solver Implementation Details}
\label{app:solver-impl}

This section lists the hyperparameters and implementation details used to instantiate \Solver\ training in \Cref{subsec:method-solver}.
Similar to \Generator, we train the \Solver\ with GRPO using two verifiable rewards: a graded format reward $r_{\text{fmt}}$ and a dense accuracy reward $r_{\text{acc}}$.
However, different from \Generator\, our \Solver\ prompt follows a Tool-Integrated Reasoning (TIR) interface: the model emits reasoning in \thinkS\ and a predicted tool-call list in \answerS, as described in \Cref{subsec:method-solver}.

\subsection{Training Setup}
Training is performed under the same infrastructure as the \Generator: HuggingFace Accelerate with DeepSpeed ZeRO-3~\citep{rajbhandari2020zero} in mixed precision (\texttt{bfloat16}) on three GPUs.
We use a per-device batch size of $2$ with gradient accumulation over $5$ steps, yielding a slightly larger global batch size of $32$ sequences per update compared to the \Generator's $24$.
This increase reflects the higher variance in \Solver\ rollouts due to the open-ended nature of tool-call prediction, where larger batches help stabilize advantage estimation across diverse task difficulties.
The optimizer is AdamW~\citep{loshchilov2017adamw} with a fixed learning rate of $1\times10^{-6}$ and weight decay of $1\times10^{-2}$, matching the \Generator\ configuration.
We train for $50$ steps per self-play iteration on the curated dataset of $2{,}000$ samples constructed through the pipeline described in \Cref{subsec:dataset-construction}, with tasks ordered from easy to hard based on the \Solver's own answer consistency.
During GRPO updates, we sample $4$ generations per prompt at temperature $1.0$ to compute relative advantages.
The maximum completion length is set to $4096$ tokens, which accommodates the full TIR output including extended reasoning traces and multi-call tool predictions.
The KL penalty coefficient is set to $\lambda_{\mathrm{KL}} = 1\times10^{-2}$ to regularize the policy against the reference model.
All hyperparameters are summarized alongside the \Generator\ configuration in \Cref{tab:hyperparams-gen-solver}.

\subsection{Reward Hyperparameters}

\paragraph{Output interface and parsing.}
A completion is considered structurally valid if it contains a non-empty \answerS\ block.
We parse \answerS\ with a super-relaxed loader that accepts: (i) strict JSON, (ii) Python-literal style dictionaries or lists (e.g., single quotes), and (iii) code-fenced JSON.
To avoid silent parsing artifacts, we treat ellipsis-like placeholders (e.g., \texttt{"..."} or \texttt{[...]} patterns) as invalid and assign zero reward.
Parsed tool calls are normalized into a canonical schema $\{\texttt{name}, \texttt{arguments}\}$.
Normalization supports common wrappers (e.g., OpenAI-style \texttt{"function":\{...\}}) and converts singleton dict outputs into a length-one list.
When arguments appear outside an explicit \texttt{"arguments"} field, we fall back to a flat argument map for robustness.

\paragraph{Format reward ($r_{\text{fmt}}$).}
We use the graded parseability reward defined in \Cref{subsec:method-solver}:
\begin{equation}
r_{\text{fmt}}(\hat{y})=
0.3\cdot\mathbb{I}_{\text{tag}}
+0.3\cdot\mathbb{I}_{\text{parse}}
+0.4\cdot\mathbb{I}_{\text{norm}},
\end{equation}
where $\mathbb{I}_{\text{tag}}$ checks presence of \answerS, $\mathbb{I}_{\text{parse}}$ checks that the enclosed content parses under the relaxed loader, and $\mathbb{I}_{\text{norm}}$ checks that normalization yields at least one canonical tool call.
This reward stabilizes early training by providing non-zero signal before full functional correctness becomes common.

\paragraph{Accuracy reward ($r_{\text{acc}}$).}
For function-call correctness, we compute a dense soft-matching reward between predicted calls $\hat{C}$ and gold calls $C^\star$.
As described in \Cref{subsec:method-solver}, each gold call is greedily matched to the highest-scoring unused prediction.
For a matched pair $(\hat{c},c^\star)$, we compute:
(i) exact tool-name match $s_{\text{name}}\in\{0,1\}$,
(ii) argument-key overlap $s_{\text{key}}\in[0,1]$ as F1 over key sets,
and (iii) value match $s_{\text{val}}\in[0,1]$ as the fraction of matching values over intersecting keys.
The per-pair score is a convex combination:
\begin{equation}
s(\hat{c},c^\star)
=\lambda_{\text{name}}s_{\text{name}}
+\lambda_{\text{key}}s_{\text{key}}
+\lambda_{\text{val}}s_{\text{val}},
\end{equation}
with $(\lambda_{\text{name}},\lambda_{\text{key}},\lambda_{\text{val}})=(0.2,\,0.3,\,0.5)$.
We average the matched scores over gold calls to obtain a base accuracy score $\bar{s}$.

\paragraph{Robust value comparison.}
Value matching uses a conservative comparator to reduce sensitivity to superficial formatting.
We treat two values as equal if they match exactly, or if they match after numeric coercion and whitespace normalization.
To avoid float precision artifacts, long numeric strings are treated as identifiers and compared as normalized strings rather than coerced floats.
If neither numeric nor string normalization applies, we fall back to canonical JSON comparison.

\paragraph{Penalty for extra tool calls.}
To discourage over-prediction and tool-call spamming, we apply a multiplicative penalty for extra predicted calls:
\begin{equation}
r_{\text{acc}} = \bar{s}\cdot \frac{1}{1+\alpha\cdot \max(0,|\hat{C}|-|C^\star|)},
\end{equation}
with $\alpha=0.25$ (\texttt{EXTRA\_CALL\_PENALTY\_ALPHA}).
This penalty leaves correct-length predictions unchanged while downweighting completions that append spurious calls.

\Cref{tab:solver-reward-hyperparams} summarizes the fixed reward-related hyperparameters used for \Solver\ training.

\begin{table}[t]
\centering
\small
\begin{tabular}{lc}
\toprule
\textbf{Hyperparameter} & \textbf{Value} \\
\midrule
Accuracy weights $(\lambda_{\text{name}},\lambda_{\text{key}},\lambda_{\text{val}})$ & $(0.2,\,0.3,\,0.5)$ \\
Extra-call penalty coefficient $\alpha$ & $0.25$ \\
Format reward weights $(\mathbb{I}_{\text{tag}},\mathbb{I}_{\text{parse}},\mathbb{I}_{\text{norm}})$ & $(0.3,\,0.3,\,0.4)$ \\
\bottomrule
\end{tabular}
\caption{\textbf{Reward hyperparameters for \Solver\ training.}}
\label{tab:solver-reward-hyperparams}
\end{table}

\section{Further Details on Evaluation} 
\label{app:evaluation}
We conduct a comprehensive evaluation across different agentic tool-calling tasks to collectively assess diverse aspects of function invocation. 
Tool-Alpaca~\citep{tang2023toolalpaca} examines generalization across heterogeneous tool categories, emphasizing robustness to synthetic distribution shifts. 
Seal-Tools~\citep{wu2024sealtool} extends this setting to large-scale APIs spanning diverse domains, reducing potential data contamination while stressing scalability. 
NexusRaven~\citep{srinivasan2023nexusraven} focuses on high-fidelity function execution over realistic APIs drawn from enterprise and cybersecurity domains, where precise adherence to function signatures is essential. 
Differently, API-Bank~\citep{li2023apibank} evaluates multi-turn scenarios that require models to select appropriate APIs within conversational context.
We additionally include SNIPS~\cite{coucke2018snips}, a spoken language understanding dataset that we adapt to function-calling format, introducing natural language variation absent from synthetic benchmarks.
All benchmarks are evaluated using Abstract Syntax Tree matching metric, which verifies structural correctness of function names, parameter presence, and type adherence.
\section{Training Details of Supervised Baseline Tool-Calling Agents}
\label{app:baselines}

To contextualize the zero-data effectiveness of \toolrzero, we compare against several prominent tool-calling agents that were originally trained on comprehensive, curated supervised datasets. Below, we briefly describe each baseline:
\begin{itemize}[topsep=-1.5pt, leftmargin=10pt, itemsep=-0pt]
    \item \textbf{xLAM~\citep{zhang2025xlam}:} xLAM is a family of Large Action Models designed to empower AI agent systems through enhanced function-calling capabilities. It aggregates diverse agent trajectories from various environments and is originally fine-tuned on a unified dataset of approximately 60,000 samples from sources like ToolBench, Webshop, ToolAlpaca, HotpotQA, AlfWorld, APIBank, Mind2Web, AgentBoard, AgentBench, and synthetic datasets such as API-GEN and SpecTools, using supervised fine-tuning (SFT) on models ranging from 1B to 8x22B parameters. For trainings, we used official data provided by authors at Huggingface: \url{https://huggingface.co/datasets/Salesforce/xlam-function-calling-60k}.
    \item \textbf{Hammer~\citep{lin2025hammer}:} It focuses on robust function-calling for on-device language models via function masking techniques to improve generalization and reduce overfitting. It is originally trained on the augmented xLAM-function-calling-60k dataset that includes around 210,000 samples and 7,500 irrelevance detection samples, using SFT on Qwen 2.0 series models (0.5B to 7B parameters), emphasizing advanced training methods over data refinement. For generating the main training dataset, we used the official codebase provided by the authors and run it for the corresponding data generation resulting in 210,000 samples: \url{https://github.com/MadeAgents/Hammer}.
    \item \textbf{ToolACE~\citep{liu2025toolace}:} An automatic agentic pipeline that generates accurate, multi-turn, and diverse tool-learning data through self-evolution synthesis and multi-agent interactions, creating over 500,000 dialogs based on 26,507 diverse APIs. It employs SFT on Llama-3.1-8B-Instruct, with a dual-layer verification system (rule-based and model-based) to ensure data quality, supporting single, parallel, nested calls, and multi-turn dialogues. We used the publicly available dataset from Huggingface: \url{https://huggingface.co/datasets/Team-ACE/ToolACE}.
    \item \textbf{ToolRL~\citep{qian2025toolrl}:} A RL framework for tool learning, arguing that reward signals are sufficient without heavy reliance on SFT data curation. It is originally trained on a mixed dataset of 4,000 samples (2,000 from ToolACE, 1,000 from masked Hammer, and 1,000 from xLAM) using GRPO, focusing on reward design for multi-step interactions, irrelevant tool detection, and generalization across unseen scenarios. For trainings we use the official code, main scripts and datasets provided by authors: \url{https://github.com/qiancheng0/ToolRL}.
\end{itemize}
Since none of these methods release model checkpoints trained on identical base models, a direct comparison of published results would conflate differences in base model capabilities with differences in training data and methodology. To ensure a controlled and fair evaluation, we re-train all baselines on same base model Qwen-2.5-1.5B-Instruct, using the officially released datasets provided by each method's respective authors.

For the three SFT-based baselines (xLAM, Hammer, and ToolACE), we conduct supervised fine-tuning using LLaMA-Factory~\citep{zheng2024llamafactory}, following the original training configurations as closely as possible (e.g., learning rate schedules, number of epochs, and context length). For ToolRL, we use the official training repository and codebase released by the authors. We train using GRPO with the same Qwen-2.5-1.5B-Instruct, adopting the default hyperparameters specified by the authors main scirpts.

In total, we try to re-implement each baseline faithfully (SFT methods with SFT, and RL methods with RL) on Qwen-2.5-1.5B-Instruct model, using author-provided datasets and closely matched training procedures. This unified setup isolates the effect of training data and algorithmic design from confounding factors such as base model choice, enabling a rigorous assessment of \toolrzero's zero-data effectiveness relative to these data-intensive approaches.

\section{Full Algorithm of \toolrzero}
\label{app:algortihm}

\Cref{alg:toolrzero} presents the complete pseudocode of the \toolrzero\ self-evolution framework.
The algorithm takes a base LLM and a task-specification distribution as input and returns a trained \Solver\ policy after $K$ co-evolutionary iterations.
Each iteration proceeds through three color-coded stages: \textcolor{acadMustard}{\textbf{Generator Training}} (yellow), where the \Generator\ learns to synthesize challenging tasks guided by the frozen \Solver's competence frontier; \textcolor{acadGreen}{\textbf{Dataset Construction}} (green), where generated tasks are deduplicated, cross-verified, and organized into a difficulty-based curriculum; and \textcolor{acadNavy}{\textbf{Solver Training}} (blue), where the \Solver\ is trained on this curated curriculum with dense accuracy rewards.

\begin{algorithm}[h]
\caption{\toolrzero: Zero-Data Self-Play for Tool-Calling Agents}
\label{alg:toolrzero}
\begin{algorithmic}[1]
\Require Base LLM $\pi$; iterations $K$; task-spec distribution $p(s)$ over $s{=}(d,c,m,n)$; probe rollouts $M$; band $[P_{\text{low}},P_{\text{high}}]$; width $\sigma$
\State Initialize \Generator\ $\pi_\theta^{(0)} \leftarrow \pi$, \Solver\ $\pi_\phi^{(0)} \leftarrow \pi$

\For{$t = 1,\ldots,K$}

  \AlgSection{acadMustardBg}{\textbf{\textcolor{acadMustard}{\textbf{Generator Training}} (\Cref{subsec:method-generator}): train \Generator\ with frozen \Solver}
  \State Freeze \Solver\ $\pi_\phi^{(t-1)}$
  \For{$u=1,\ldots,U_G$} \Comment{GRPO steps}
    \State Sample task specification $s \sim p(s)$
    \State Sample $x \sim \pi_\theta^{(t-1)}(\cdot \mid s)$ yielding $(q,\mathcal{T},c^\star)$
    \State Compute $r_{\text{fmt}}(x)$ \Comment{Eq.~(\ref{eq:1})}
    \State Compute $r_{\text{valid}}(x)$ \Comment{Eq.~(\ref{eq:2})}
    \State Probe frozen \Solver\ $M$ times; estimate $\hat{p}_{\text{succ}}$
    \State Compute $r_{\text{diff}}(x)$ (band-pass on $\hat{p}_{\text{succ}}$) \Comment{Eq.~(\ref{eq:r_diff})}
    \State Compute $r_{\text{sem}}(x)$ (semantic alignment scoring)
    \State $r_{\text{curr}}(x)\leftarrow r_{\text{diff}}(x)+r_{\text{sem}}(x)$ \Comment{Eq.~(\ref{eq:4})}
    \State $R_G(x)\leftarrow r_{\text{fmt}}(x)+r_{\text{valid}}(x)+r_{\text{curr}}(x)$
  \EndFor
  }
  \State $\pi_\theta^{(t)} \leftarrow \textsc{GRPO}\!\left(\pi_\theta^{(t-1)}, R_G\right)$

  \AlgSection{acadGreenBg}{\textbf{\textcolor{acadGreen}{\textbf{Dataset Construction}} (\Cref{subsec:dataset-construction}): curate curriculum from frozen \Generator}
  \State Freeze \Generator\ $\pi_\theta^{(t)}$
  \State Sample candidate pool $\mathcal{P}$ from $\pi_\theta^{(t)}(\cdot \mid s)$
  \State Deduplicate via canonicalized question--tool--call signatures
  \State Cross-verify with frozen \Solver\ $\pi_\phi^{(t-1)}$; retain consistent tasks
  \State Estimate difficulty (pass@$M$) and bucket into \{easy, medium, hard\}
  \State Construct curriculum $\mathcal{D}_t$: domain-balanced mix, ordered easy $\rightarrow$ hard
  }

  \AlgSection{acadNavyBg}{\textbf{\textcolor{acadNavy}{\textbf{Solver Training}} (\Cref{subsec:method-solver}): train \Solver\ on curated curriculum}
  \For{$u=1,\ldots,U_S$} \Comment{GRPO steps}
    \State Sample minibatch $B \sim \mathcal{D}_t$ (easy $\rightarrow$ hard)
    \State Sample $\hat{y}\sim \pi_\phi^{(t-1)}(\cdot \mid q,\mathcal{T})$ yielding predicted calls $\hat{C}$
    \State Compute $r_{\text{fmt}}(\hat{y})$ (tag/parse/normalization)
    \State Compute $r_{\text{acc}}(\hat{C},C^\star)$ (name/key/value + extra-call penalty)
    \State $R_S(\hat{y}) \leftarrow r_{\text{fmt}}(\hat{y}) + r_{\text{acc}}(\hat{C},C^\star)$
  \EndFor
  }
  \State $\pi_\phi^{(t)} \leftarrow \textsc{GRPO}\!\left(\pi_\phi^{(t-1)}, R_S\right)$

\EndFor
\State \Return $\pi_\phi^{(K)}$
\end{algorithmic}
\end{algorithm}

\newpage
\clearpage
\begin{figure*}[t]
\centering

\begin{tcolorbox}[
  enhanced,
  width=0.98\linewidth,
  colback=tzBlueFill,
  colframe=tzBlueBorder,
  boxrule=1.2pt,
  arc=6pt,
  left=5pt,right=5pt,top=4pt,bottom=2pt,
  title=\textbf{\large Generator Prompt},
  coltitle=white,
  colbacktitle=tzBlueHeader2,
  fonttitle=\bfseries,
]
\small
\begin{lstlisting}[style=jsonTiny]
You are an expert task generator for tool-calling agents.

FIRST, in your private scratch-pad, reason step-by-step to design a realistic, non-trivial task that cannot be solved without correctly calling one or sometimes multiple tools.

CONTROL SPEC (MUST FOLLOW EXACTLY):
- Domain: {domain}
- Context type: {context_type}  (single_turn or multi_turn)
- Number of available tools: {tool_menu_size} (<available_tools>)
- Number of gold tool calls: {num_calls} (<tool_call_answer>)

RULES TO SATISFY THE SPEC:
1) You MUST output exactly {tool_menu_size} tools in <available_tools>.
2) You MUST output exactly {num_calls} tool calls (JSON list length) in <tool_call_answer>.
3) Domain must be {domain}. Do not drift into other domains.
4) If context_type=multi_turn, embed a short conversation in <question> like: "# Conversation\nUser: ...\nAgent: ...\nUser: ...\nAgent: ..."
5) Tool arguments must be flat primitives only (no lists, no nested objects).
6) The function values (<value1>, <value2>, ...) MUST be present inside user question (<question>...</question>), otherwise agent cannot solve the task.

THEN, without revealing your reasoning, output the following four blocks in the exact format, NOTHING ELSE:

<think>
Your private reasoning here.
</think>

<question>
Write a natural user question (no bullet points, no meta-instructions, no placeholders).
It must be a natural question, be in domain "{domain}", and mention the exact argument values that appear in <tool_call_answer>.
</question>

<available_tools>
A JSON list of tools. Each tool MUST include: "name", "description", and "parameters". 
[
    {
        "name": "<tool_name>",
        "description": "<short description>",
        "parameters": {
        "<param1>": {"type": "<param1_type>", "description": "<param1_description>"},
        "<param2>": {"type": "<param2_type>", "description": "<param2_description>"},
        ...
        },
        "required": [<param1>, ...],
    },
    ...
]
</available_tools>

<tool_call_answer>
[
{\"name\": \"<tool_name>\", \"arguments\": {\"<param>\": <value>, ...}}
]
</tool_call_answer>

Generate a new tool-calling task now. Follow the CONTROL SPEC exactly and remember to format the output exactly as instructed.
\end{lstlisting}

\end{tcolorbox}

\caption{\textbf{\Generator\ prompt used for task synthesis in \toolrzero.} The system template enforces domain/context constraints, tool-menu cardinality, and primitive-only arguments, while requiring the final output to follow a strict four-block schema (\thinkS, \questionS, \menuS, \answerS).}
\label{fig:generator-prompt-template}
\end{figure*}

\newpage
\clearpage
\begin{figure*}[t]
\centering

\begin{tcolorbox}[
  enhanced,
  width=0.98\linewidth,
  colback=tzBlueFill,
  colframe=tzBlueBorder,
  boxrule=1.2pt,
  arc=6pt,
  left=5pt,right=5pt,top=4pt,bottom=2pt,
  title=\textbf{\large Solver Prompt},
  coltitle=white,
  colbacktitle=tzBlueHeader2,
  fonttitle=\bfseries,
]
\small
\begin{lstlisting}[style=jsonTiny]
You are a strict quality control judge for a synthetic data generation pipeline.
You will be given a User Question, Available Tools, and a Tool Call Answer.

Your job is to score the example on a scale of 1 to 5 based on TWO criteria:
1. **Question Quality**: Is the user question realistic, specific, and clear? (CRITICAL)
2. **Semantic Coherence**: Does the tool call actually solve the user's request?

Scoring Rubric:
- 5 (Perfect): The question is specific and realistic (e.g., "Book a flight to Paris on Dec 5th"). The tool call perfectly addresses it.
- 4 (Good): The question is good, but the tool call has minor issues (e.g., slightly different parameter values that still work).
- 3 (Passable): The question is vague or simple. The tool call matches it.
- 2 (Bad Question): The question is generic, placeholder text (e.g., "User request here", "Make a tool call"), or nonsense. **Score 2 or 1 immediately if the question is bad.**
- 1 (Failure): The tool call is completely unrelated, OR the question is clearly a template error (e.g., "A single concrete user request").

**IMPORTANT:** If the User Question looks like an instruction (e.g., "Generate a query...") rather than a natural user request, you MUST give a score of 1 or 2.

Reply with ONLY a single integer from 1 to 5.
\end{lstlisting}

\end{tcolorbox}

\caption{
\textbf{Semantic alignment reward prompt template.}
Prompt used to compute the semantic alignment reward $r_{\text{sem}}$ in \Cref{eq:4}, evaluating whether the synthesized user question is realistic and well-formed, and whether the corresponding tool call semantically and functionally satisfies the user’s request.
}

\label{fig:semantic-alignment-prompt-template}
\end{figure*}

\newpage
\clearpage
\begin{figure*}[t]
\centering

\begin{tcolorbox}[
  enhanced,
  width=0.98\linewidth,
  colback=tzBlueFill,
  colframe=tzBlueBorder,
  boxrule=1.2pt,
  arc=6pt,
  left=5pt,right=5pt,top=4pt,bottom=2pt,
  title=\textbf{\large Solver Prompt},
  coltitle=white,
  colbacktitle=tzBlueHeader2,
  fonttitle=\bfseries,
]
\small
\begin{lstlisting}[style=jsonTiny]
A conversation between user and tool-calling assistant. The user asks a question, and the assistant uses tools to solve it. The assistant first thinks about the reasoning process in the mind and then provides the user with the answer. The reasoning process and answer are enclosed within <think>...</think> and 
<tool_call_answer>...</tool_call_answer> tags, i.e.:
<think> 
This is my reasoning. 
</think>
<tool_call_answer>
[
  {
    "name": "tool_name", 
    "arguments": {"arg1": "value", "arg2": "value2", ...}
  }
]
</tool_call_answer>

User Query:
<question>
USER_QUERY
</question>

Available Tools (JSON):
<available_tools>
TOOL_MENU
</available_tools>
\end{lstlisting}

\end{tcolorbox}

\caption{
\textbf{\Solver\ prompt template.}
\textsc{\textbf{User\_Query}} indicates the user’s natural-language input,
while \textsc{\textbf{Tool\_Menu}} denotes the set of available tools;
both placeholders are instantiated dynamically at each training step.
}
\label{fig:solver-prompt-template}
\end{figure*}

\newpage
\clearpage
\begin{figure*}[t]
\centering

\begin{tcolorbox}[
  enhanced,
  width=0.98\linewidth,
  colback=tzBlueFill,
  colframe=tzBlueBorder,
  boxrule=1.2pt,
  arc=6pt,
  left=5pt,right=5pt,top=4pt,bottom=2pt,
  title=\textbf{\large Early Training (Iter 1): Surface-level User Request},
  coltitle=white,
  colbacktitle=tzBlueHeader2,
  fonttitle=\bfseries,
]
\small \textbf{User Request:} Book a flight from London to Paris.

\medskip
\textbf{Available tools (JSON):}
\begin{lstlisting}[style=jsonTiny]
[
  {
    "name": "FlightBookingService",
    "description": "A tool to book flights using a travel service.",
    "parameters": {
      "type": "object",
      "properties": {
        "origin": {"type": "string", "description": "Origin airport name"},
        "destination": {"type": "string", "description": "Destination airport name"}
      },
      "required": [origin, destination]
    }
  }
]
\end{lstlisting}

\textbf{Output}: 
\begin{lstlisting}[style=jsonTiny]
[
  {
    "name": "FlightBookingService", 
    "arguments": {"origin": "LHR", "destination": "CDG"}
  }
]
\end{lstlisting}
\end{tcolorbox}
\caption{\textbf{Early-stage \Generator\ behavior in \toolrzero.} In the first training iteration, the \Generator\ produces surface-level, minimally structured tool-use tasks. The user request is simple and underspecified, the tool menu contains a single available function, and the solution consists of a single, straightforward tool call with canonical arguments.}
\label{fig:gen-evolution-travel-easy}
\end{figure*}

\newpage
\clearpage
\begin{figure*}[t]
\centering
\begin{tcolorbox}[
  enhanced,
  width=0.98\linewidth,
  colback=tzBlueFill,
  colframe=tzBlueBorder,
  boxrule=1.2pt,
  arc=6pt,
  left=5pt,right=5pt,top=4pt,bottom=2pt,
  title=\textbf{\large Late Training (Iter 3): Complex User Request with Multi-step Tool-Calling},
  coltitle=white,
  colbacktitle=tzBlueHeader2,
  fonttitle=\bfseries,
]
\small \textbf{User request:} What is the best way to book a round-trip from \textbf{New York} to \textbf{Paris} in \textbf{first week of July 2023} for \textbf{two passengers} at \textbf{business class}, and also \textbf{book a hotel in central Paris} for \textbf{one night}?

\medskip
\textbf{Available tools (JSON):}
\begin{lstlisting}[style=jsonTiny]
[
  {
    "name": "FlightBooking",
    "description": "A service to book flights.",
    "parameters": {
      "type": "object",
      "properties": {
        "departureCity": {"type": "string", "description": "The city where the flight is departing."},
        "arrivalCity": {"type": "string", "description": "The city where the flight is arriving."},
        "departureDate": {"type": "string", "description": "The date of departure for the round-trip flight."},
        "arrivalDate": {"type": "string", "description": "The date of arrival for the round-trip flight."},
        "passengerCount": {"type": "integer", "description": "The number of passengers for the round-trip flight."},
        "cabinClass": {"type": "string","description": "Cabin class (e.g., economy, premium_economy, business, first)."}
      },
      "required": ["departureCity", "arrivalCity", "departureDate", "arrivalDate", "passengerCount", "cabinClass"]
    }
  },
  {
    "name": "HotelBooking",
    "description": "A service to book hotels.",
    "parameters": {
      "type": "object",
      "properties": {
        "city": {"type": "string","description": "City where the hotel is located."},
        "checkInDate": {"type": "string", "description": "Hotel check-in date."},
        "checkOutDate": {"type": "string", "description": "Hotel check-out date."},
        "guestCount": {"type": "integer", "description": "Number of guests."},
        "locationPreference": {"type": "string", "description": "Preferred area/neighborhood (e.g., 'central Paris', 'city center')."
        }
      },
      "required": ["city", "checkInDate", "checkOutDate", "guestCount", "locationPreference"]
    }
  }
]
\end{lstlisting}

\textbf{Answer:}
\begin{lstlisting}[style=jsonTiny]
[
  {
    "name": "FlightBooking",
    "arguments": {"departureCity": "New York", "arrivalCity": "Paris", "departureDate": "2023-06-01", "arrivalDate": "2023-06-02", "passengerCount": 2, "cabinClass": "business"}
  },
  {
    "name": "HotelBooking",
    "arguments": {"city": "Paris", "checkInDate": "2023-06-02", "checkOutDate": "2023-06-03", "guestCount": 2, "locationPreference": "central"}
  }
]
\end{lstlisting}
\end{tcolorbox}

\caption{\textbf{Late-stage \Generator\ behavior in \toolrzero.} By Iteration 3, the \Generator\ synthesizes complex, multi-constraint user requests that require multi-step tool execution. The task jointly involves booking a round-trip flight and reserving a centrally located hotel, incorporating temporal constraints, passenger count, and cabin class preferences. The generated solution correctly decomposes the request into multiple ordered tool calls, demonstrating improved task abstraction, constraint integration, and compositional planning.}
\label{fig:gen-evolution-travel-hard}
\end{figure*}

\end{document}